\documentclass[letterpaper,journal]{IEEEtran}
\usepackage{amsmath,amssymb,amsfonts}
\usepackage{algorithmic}
\usepackage{array}
\usepackage[caption=false,font=normalsize,labelfont=sf,textfont=sf]{subfig}
\usepackage{textcomp}
\usepackage{url}
\usepackage{verbatim}
\usepackage{graphicx}
\usepackage{cite}
\usepackage{xcolor}
\usepackage{booktabs}
\usepackage{multirow}
\usepackage{makecell}
\usepackage{hyperref}

\begin{document}

\title{FDIM: A Feature-distance-based Generic Video Quality Metric for Versatile Codecs}

\author{
Jiayi~Wang,
Lichun Zhang, 
Xiaoqi~Zhuang,
Jiaqi~Zhang,
Lu~Yu,
and~Yin Zhao
\thanks{J. Wang, X. Zhuang, J. Zhang, and L. Yu are with the College of Information Science and Electronic Engineering, Zhejiang University, and also with the Zhejiang Key Laboratory of Multimodal Communication Networks and Intelligent Information Processing, Hangzhou 310058, China (e-mail: \{blindwang, zhuangxiaoqi, jiaqi.zhang, yul\}@zju.edu.cn).}%
\thanks{L. Zhang, Y. Zhao are with the Central Media Technology Institute, Huawei, Hangzhou 310058, China (e-mail: \{zhanglichun5, yin.zhao\}@huawei.com).}%
\thanks{Corresponding author: Lu Yu.}%
}

\maketitle

\begin{abstract}
Video technology is advancing toward Ultra High Definition (UHD) and High Dynamic Range (HDR), which intensifies the need for higher compression efficiency for these high-specification videos. Beyond advances in traditional codecs, neural video codecs (NVCs) have attracted significant research attention and have evolved rapidly over the past few years. The coding artifacts of NVCs often exhibit content-varying and generative characteristics, which differ from those of conventional codecs and are challenging for traditional video quality assessment (VQA) methods to capture.
Therefore, VQA metrics are required to generalize across different codecs, content types, and dynamic ranges to better support video codec research and evaluation.
In this paper, we propose FDIM, a feature-distance-based generic video quality metric for both traditional and neural video codecs across SDR and HDR formats.
FDIM employs a hybrid architecture that integrates deep and hand-crafted features. The deep feature component learns multi-scale representations to capture distortions ranging from structural and textural fidelity degradation to high-level semantic deviations, while the hand-crafted feature component provides stable complementary cues to improve overall generalization.
Specifically, the deep feature component employs content-adaptive feature-distance modeling and attention-based multi-scale fusion, enabling perceptual difference estimation aligned with human visual sensitivity patterns.
We trained FDIM on a large-scale subjective quality assessment dataset (DCVQA) consisting of over 16k video sequences encoded by traditional block-based hybrid video codecs and end-to-end perceptually optimized neural video codecs. 
Extensive experiments on ten SDR/HDR VQA datasets containing diverse, previously unseen codecs demonstrate that FDIM achieves strong generalization and high correlation with subjective assessment.
The source code for FDIM and the DCVQA validation set will be released at \url{https://github.com/MCL-ZJU/FDIM}.
\end{abstract}


\section{Introduction}
\IEEEPARstart{V}{ideo} delivery is rapidly shifting towards Ultra High Definition (UHD), High Dynamic Range (HDR), and Wider Color Gamut (WCG) to meet the demand for more immersive and realistic viewing experiences. This trend intensifies the need for efficient compression of UHD and HDR videos with high visual quality~\cite{sullivan2012overview,bross2021overview}. Neural video coding has emerged as a promising direction for further improving compression efficiency, ranging from neural-enhanced hybrid codecs~\cite{NNVC} to end-to-end codecs~\cite{lu2019dvc,hu2021fvc,shi2022alphavc,Jia_2025_CVPR}. In practical deployment, quality metrics must be generic and remain reliable across heterogeneous delivery conditions and content pipelines, where distortions vary substantially across codecs, content types, and dynamic ranges. Therefore, metrics trained or calibrated on a limited set of codecs or formats often exhibit limited generalizability.

\begin{figure*}[t]
\centering
\includegraphics[width=1\textwidth]{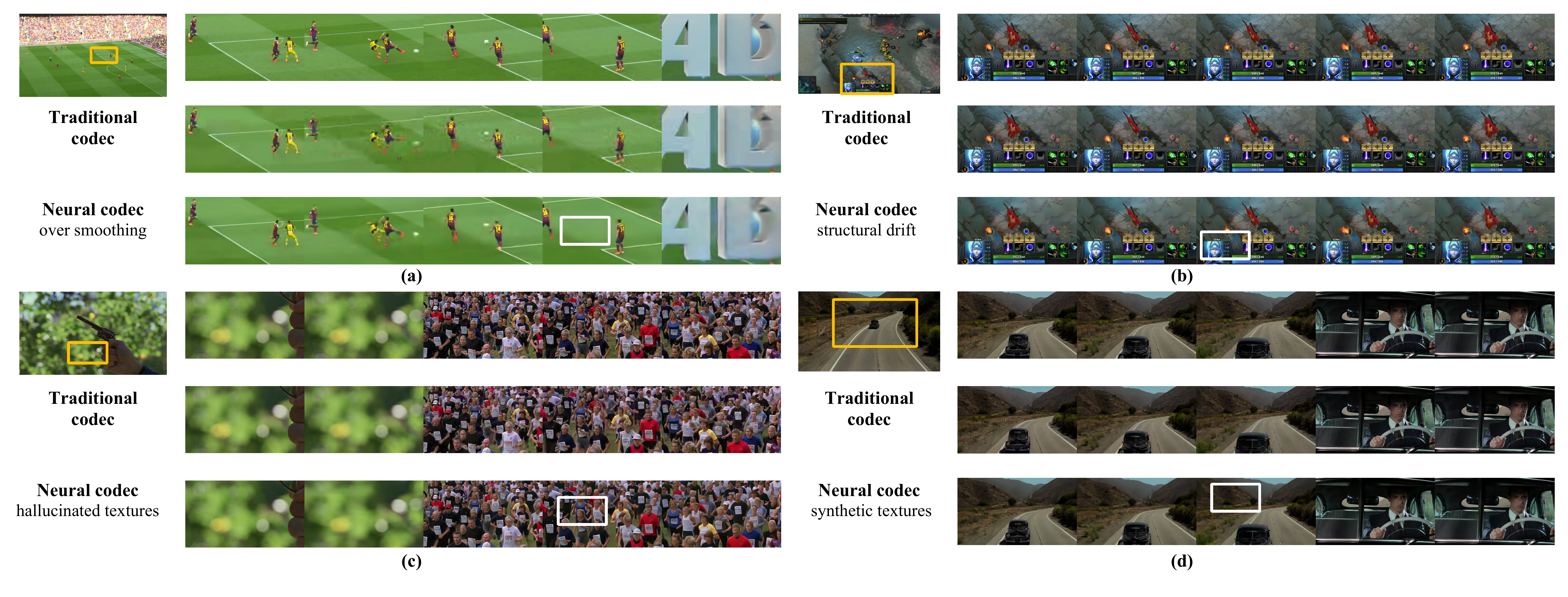}
\caption{Illustration of characteristic artifacts introduced by neural video codecs in comparison with traditional codecs. 
For each sequence, frames are uniformly sampled at a 30-frame interval. 
Yellow boxes indicate the cropped regions that are enlarged for visualization, while white boxes highlight representative artifact patterns. }
\label{fig:artifact}
\end{figure*}

The human visual system is inherently hierarchical~\cite{Deep_Hierarchies_TPAMI}, progressing from low-level perceptual encoding in early visual areas to high-level cognitive processing in higher cortical regions, which supports semantic understanding and contextual reasoning.
Traditional quality metrics~\cite{wang2004image,sheikh2006image,zhang2011fsim} are closely aligned with low-level perceptual processing and effective for classical structured artifacts such as ringing, blocking, and blur, which typically manifest as predictable changes in low-level signal statistics. 
However, NVCs tend to introduce content-varying and generative distortions that differ fundamentally from the structured artifacts produced by conventional codecs. As illustrated in Fig.~\ref{fig:artifact}, neural reconstruction may produce over-smoothing, local structural drift, and hallucinated textures. Hand-crafted features exhibit limited sensitivity to semantic similarity and perceptual quality~\cite{Blau_2018_CVPR}, highlighting their limitations in capturing the complex, content-dependent distortions introduced by NVCs.  

In contrast, deep learning-based approaches can be viewed as approximating high-level cognitive processes. 
Through large-scale data-driven training, deep neural networks implicitly learn the mapping between distorted visual stimuli and human subjective judgments. This process can be interpreted as constructing a form of perceptual cognition, where the model captures prior knowledge about distortion types, content dependency, and perceptual sensitivity. 
Deep learning-based metrics leverage backbones ranging from CNNs to transformers~\cite{kim2017deep,kim2018deep,xu2020c3dvqa,lao2022attentions,starvqa,wu2022discovqa,chen2024topiq} to learn high-level visual representations and capture content-dependent distortions. 
Despite these advances, several fundamental limitations remain. First, the strong reliance on large-scale training datasets and the high computational complexity of modern architectures hinder practical deployment and out-of-distribution generalization. Second, most existing models are primarily trained on distortions produced by conventional codecs, which limits their ability to handle the visually distinctive and generative artifacts introduced by NVCs.

Furthermore, another important challenge lies in SDR-to-HDR cross-domain generalization, where HDR subjective databases are typically smaller and exhibit signal statistics that differ substantially from SDR.
Existing SDR-trained models often lack cross-domain transferability to HDR scenarios, whereas HDR-trained models~\cite{LIVEHDR2024TIP,hdrvdp32023arXiv, colorvideovdp} are prone to overfitting due to limited dataset scale. 
Therefore, it is desirable to leverage large-scale SDR VQA data to learn stable perceptual representations, and to develop unified models that robustly cover both SDR and HDR scenarios. 

Toward a generic full-reference VQA metric applicable to versatile codecs and enabling joint modeling of hierarchical visual perception and cognition, we propose FDIM, a hybrid framework that integrates a deep feature component with a hand-crafted feature component. The deep feature component performs content-adaptive feature-distance modeling between reference and distorted features and aggregates multi-scale cues via an attention mechanism to capture semantic and generative artifacts across codecs. 
The hand-crafted feature component uses VMAF features, which we find to be complementary to the deep representations in extensive experiments.
The quality scores from the two components are mapped to a unified quality scale and fused to obtain the final prediction.
To enable HDR evaluation without training on HDR compressed videos, we adopt PU21 encoding~\cite{pu21} as a lightweight preprocessing that enables zero-shot transfer from an SDR-trained pipeline to HDR datasets.

In summary, our contributions can be summarized as follows: 
\begin{enumerate}

    \item We propose FDIM, a hybrid full-reference VQA metric that achieves strong generalization by combining complementary deep and hand-crafted features and by leveraging diverse training data.
    Specifically, FDIM jointly models hierarchical visual perception and cognition using both hand-crafted cues and learned deep representations.
    The deep branch performs content-adaptive feature-distance modeling with attention-based multi-scale fusion, capturing content-dependent perceptual discrepancies, while the hand-crafted branch provides stable and complementary cues for conventional distortions. 
    To ensure the model generalizes robustly across versatile codecs and distortion types, the model is trained on the DCVQA dataset, comprising over 16k sequences encoded by both traditional block-based codecs and end-to-end perceptually optimized neural video codecs.

    \item We design FDIM to be lightweight while maintaining high correlation with subjective quality. 
    To effectively exploit the training data, we adopt ranking-inspired learning over both identical and distinct source contents, which improves quality discriminability and generalization.
    Additionally, FDIM is designed with a moderate capacity to match the available training data, avoiding unnecessary complexity while preserving sufficient representation capability.
    Extensive experiments demonstrate that FDIM achieves a favorable trade-off between accuracy and computational efficiency.
\end{enumerate}

\section{Related Work and Challenges}
\label{sec:related}

\subsection{Traditional Quality Metrics}

Early quality metrics rely on pixel-wise fidelity, such as MSE and PSNR. Despite their simplicity and low computational complexity, these metrics often correlate poorly with human perception~\cite{wang2009mean}. To better model the human visual system, perceptually-grounded metrics such as SSIM~\cite{wang2004image}, MS-SSIM~\cite{wang2003multiscale}, VIF~\cite{sheikh2006image} and FSIM~\cite{zhang2011fsim} compare structural and feature-level similarities rather than raw errors. For video, temporal artifacts are explicitly modeled by spatio-temporal metrics such as VQM~\cite{pinson2004objective} and ST-MAD~\cite{seshadrinathan2010motion}.
VMAF~\cite{li2016toward} follows a fusion-based paradigm, combining multiple hand-crafted features via support vector regression to improve perceptual correlation.
Extensive efforts aim to improve accuracy by designing stronger atomic features and fusion strategies~\cite{ensemble_vmaf, evmaf, hfr_vmaf, color_vmaf}. 
FUNQUE~\cite{funque} utilizes a perceptually tuned wavelet-domain transform shared across all atom quality models, thereby achieving higher accuracy with reduced computational complexity. 
However, hand-crafted features exhibit limited sensitivity to semantic similarity and perceptual quality~\cite{Blau_2018_CVPR}. 

\subsection{Deep Learning-Based Quality Metrics}

Deep networks have been increasingly applied to data-driven quality models, leveraging diverse backbones such as ResNet~\cite{li2019quality}, 2D-CNNs~\cite{kim2017deep, kim2018deep}, 3D-CNNs~\cite{xu2020c3dvqa}, and Transformers~\cite{lao2022attentions, starvqa, wu2022discovqa, chen2024topiq}. To improve robustness and data efficiency, recent work explores ranking-based learning objectives~\cite{ma2017dipiq, feng2024rankdvqa} and contrastive learning~\cite{CONVIQT}.
TOPIQ~\cite{chen2024topiq} leverages multi-scale features and progressively propagates multi-level semantic information to low-level representations in a top-down manner.
RankDVQA~\cite{feng2024rankdvqa} is a transformer-based network that employs a ranking-inspired hybrid training strategy and a large-scale VQA training database without relying on human-provided ground-truth labels.
More recently, multi-modal large language models have been investigated for video quality assessment tasks~\cite{wu2023qalign, ge2024lmmvqaadvancingvideoquality}.

Despite recent progress in deep learning-based VQA metrics, computational efficiency and generalization remain key bottlenecks. 
First, deep learning models typically involve high computational complexity and require a careful balance between model capacity and the amount of training data to improve generalization. 
However, the scale of currently available VQA datasets is still limited compared to the increasing model capacity, which may lead to overfitting and reduced robustness. 
Consequently, lightweight architectures that can achieve competitive performance with limited-scale training datasets are highly desirable for practical deployment. 
Second, many learning-based metrics are mainly validated through intra-database cross-validation, making existing evaluation protocols insufficient to reliably assess generalization. 
Therefore, more comprehensive benchmarks covering a wider range of codecs and content types are needed to properly evaluate the generalization capability of objective metrics.

\subsection{Quality Assessment for Neural Video Codecs}

Unlike traditional codecs that rely on prediction–transform–quantization pipelines, neural video codecs (NVCs) replace key modules with data-driven neural networks. In these systems, transform coding is achieved by mapping pixels to quantized latent representations and reconstructing them through a nonlinear decoder~\cite{ma2023overview}. Moreover, many NVCs are optimized using perceptual objectives such as SSIM~\cite{wang2004image} and LPIPS~\cite{zhang2018unreasonable} rather than pure mean squared error, shifting the reconstruction objective toward perceptual and semantic similarity.
Consequently, NVCs tend to introduce content-varying and generative distortions that differ fundamentally from the structured artifacts produced by conventional codecs.

Several studies have started to investigate whether existing objective metrics remain reliable on neural video codecs. Majeedi et al.~\cite{Majeedi2023FullRV} showed that FR metrics designed for traditional codecs correlate poorly with subjective scores due to fundamentally different artifact characteristics of NVCs, and proposed the MLCVQA model and an associated dataset tailored to machine learning-based codecs. Herb et al.~\cite{herb2025evaluating} conducted a subjective and objective study on AV1~\cite{AV1}, VVC~\cite{VVC} and two DCVC variants~\cite{DCVC_FM, Jia_2025_CVPR}, and compared a broad set of full-reference, hybrid and no-reference metrics. Their results indicate that conventional metrics can still provide strong correlation with mean opinion scores for neural codecs. 
Jenadeleh et al.~\cite{jenadeleh2025jpegai} reported that a wide range of objective metrics maintain strong correlation with subjective judgments for JPEG~AI compressed images~\cite{JPEG_AI2024}, yet they exhibit systematic overestimation of quality impairments.
However, the subjective datasets used in~\cite{herb2025evaluating,jenadeleh2025jpegai} cover only a narrow range of source sequences and codecs, leaving it unclear whether these observations generalize to a broader spectrum of NVCs and diverse content. To address this gap, we conduct a more comprehensive evaluation across multiple neural codecs and source contents, and develop a metric that generalizes well across conventional and neural codecs.

\subsection{Quality Assessment for HDR Videos}

Existing approaches for HDR quality assessment can be broadly categorized into SDR-extended methods and HDR-specific modeling methods.
Extended SDR-based methods adapt conventional SDR metrics to HDR scenarios through preprocessing~\cite{pu21, LIVEHDR2024TIP, funque_hdr} or representation adaptation strategies~\cite{chubarau2024adaptingpretrainednetworksimage, Saini_2024_WACV}. PU-encoding methods~\cite{pu21} transform HDR signals into perceptually uniform spaces to reduce luminance-dependent visibility variation. 
HDR-specific methods focus on perceptual modeling of the human visual system, which explicitly account for contrast sensitivity, luminance adaptation, and masking effects~\cite{hdrvdp32023arXiv, colorvideovdp}.
However, these methods are frequently constrained by the limited scale and diversity of existing HDR datasets, making them prone to overfitting and reduced generalization to unseen content~\cite{CVQAHDR2025ICMEW}. 

From a perceptual perspective, SDR and HDR do not correspond to two independent viewing systems~\cite{mantiuk2011hdr,colorvideovdp}. 
HDR mainly extends the stimulus dynamic range and color gamut, which changes signal statistics and distortion visibility, but does not alter the underlying perceptual processing principles. 
Therefore, it is essential to leverage large-scale SDR VQA data to learn stable perceptual representations, as well as to develop unified models that can robustly handle both SDR and HDR scenarios.

\section{Proposed Method}
\label{sec:method}

\subsection{Overall Framework}

We propose a hybrid video quality assessment framework that combines learned deep representations with complementary hand-crafted features. As illustrated in Fig.~\ref{fig:overall_arch}(a), the framework comprises two parallel components: (1) a deep feature component with content-adaptive feature-distance modeling, and (2) a hand-crafted feature component based on VMAF. The two component outputs are independently mapped to a unified quality scale via nonlinear functions and then averaged to obtain the final quality score.

Mathematically, let $R$ denote the reference video and $D$ the distorted video. The hand-crafted feature component computes $q_{\text{Trad}} = \text{VMAF}(R, D)$, while the deep feature component produces $q_{\text{Deep}} = f_{\theta}(R, D)$ where $f_{\theta}$ is a neural network with parameters $\theta$. These component outputs are mapped through component-specific four-parameter logistic functions:
\begin{equation}
    \label{eq:branch_mapping}
\tilde{q}_i = \psi_i \left( q_i \right)=\beta_{i,1} \left( \frac{1}{2} - \frac{1}{1 + \exp(\beta_{i,2}(q_i - \beta_{i,3}))} \right) + \beta_{i,4}
\end{equation}
where $i \in \{\text{Trad}, \text{Deep}\}$ and each component has its own fitted parameter set $\{\beta_{i,1}, \beta_{i,2}, \beta_{i,3}, \beta_{i,4}\}$. 
The final predicted quality is:
\begin{equation}
Q = \frac{1}{2}(\tilde{q}_{\text{Deep}} + \tilde{q}_{\text{Trad}})
\end{equation}

This hybrid design leverages complementary strengths: the deep feature component extracts multi-scale representations that capture distortions ranging from structural and textural fidelity degradations to high-level semantic deviations, with particular sensitivity to the generative artifacts characteristic of NVCs; the hand-crafted feature component complements this with stable and interpretable assessments grounded in established perceptual models.

\subsection{Deep Feature Component Architecture}

\begin{figure*}[t]
\centering
\includegraphics[width=1\textwidth]{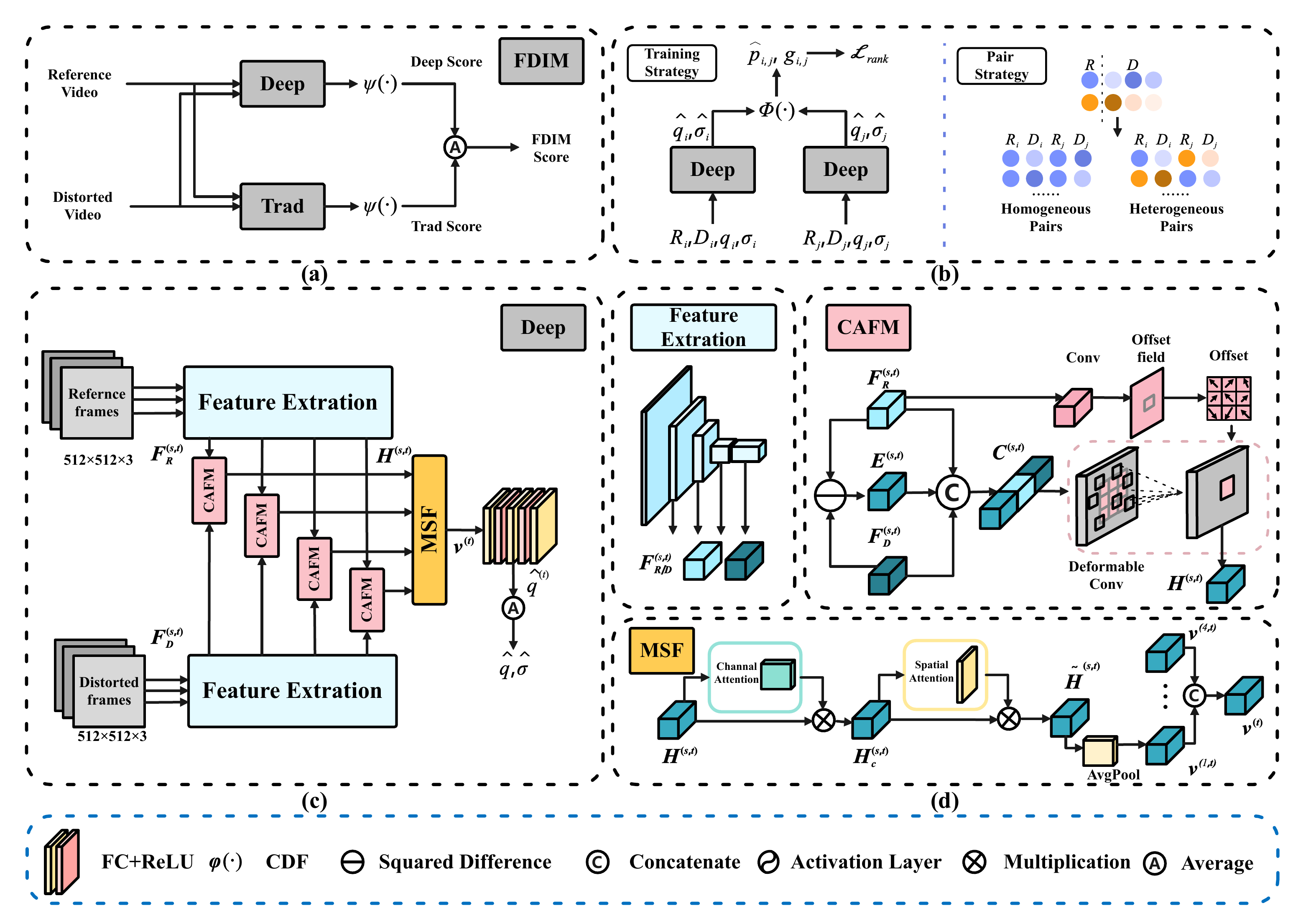}
\caption{Architecture of FDIM. The framework consists of a deep feature component and a complementary hand-crafted feature component based on VMAF features. For the deep feature component, a weight-shared ResNet-18 extracts multi-scale features from reference and distorted frames. CAFM conducts content-adaptive feature-distance modeling at each scale. The resulting features are fused via attention-based multi-scale fusion and regressed to frame-level quality scores. The outputs of the deep and hand-crafted components are finally fused on a unified quality scale to obtain the predicted score.}
\label{fig:overall_arch}
\end{figure*}

As illustrated in Fig.~\ref{fig:overall_arch}(b)–(d), the deep feature component consists of four stages: (i) Multi-Scale Feature Extraction, (ii) Content-Adaptive Feature-Distance Modeling (CAFM), (iii) Attention-based Multi-Scale Fusion (MSF), and (iv) Quality Regression.

\subsubsection{Multi-Scale Feature Extraction}

We adopt a ResNet-18 backbone~\cite{he2016deep} pretrained on ImageNet~\cite{russakovsky2015imagenet} as the feature extractor, balancing representation capacity and computational efficiency. Given a reference frame $R^{(t)}$ and a distorted frame $D^{(t)}$ at time index $t$, a weight-shared encoder extracts four-scale feature pyramids from the stages $\{\text{conv2\_x},\text{conv3\_x},\text{conv4\_x},\text{conv5\_x}\}$:
\begin{align}
\{F_R^{(s,t)}\}_{s=1}^{4} &= \text{ResNet}(R^{(t)}), \\
\{F_D^{(s,t)}\}_{s=1}^{4} &= \text{ResNet}(D^{(t)}),
\end{align}
where $s$ indexes the spatial scale (from shallow to deep). Pretraining provides robust low-level primitives, while fine-tuning adapts the representation to distortion-aware quality prediction.

\subsubsection{Content-Adaptive Feature-Distance Modeling}

The core of our deep feature component is the CAFM module, which performs content-adaptive feature-distance modeling between $F_R^{(s,t)}$ and $F_D^{(s,t)}$. 
This design is motivated by content-dependent sensitivity in human vision: distortion visibility varies with local luminance adaptation, spatial structure, and texture masking.

For each scale $s$, CAFM proceeds in three steps.

\textbf{Point-wise squared error.} We first compute an element-wise discrepancy map in feature space:
\begin{equation}
E^{(s,t)} = \big(F_R^{(s,t)} - F_D^{(s,t)}\big)^2.
\end{equation}

\textbf{Feature concatenation.} We then concatenate the reference features, distorted features, and discrepancy features along the channel dimension. The discrepancy features explicitly encode distortion magnitude, while the reference and distorted features provide contextual information to model spatial masking and modulate the perceived impact of distortions.
\begin{equation}
C^{(s,t)} = \text{Concat}\!\Big(F_R^{(s,t)},\,F_D^{(s,t)},\,E^{(s,t)}\Big).
\end{equation}

\textbf{Deformable convolution for content-adaptive aggregation.} The concatenated tensor is fed into a deformable convolution layer~\cite{dai2017deformable}, whose offsets are conditioned on the reference features:
\begin{equation}
H^{(s,t)} = \text{DeformConv}\!\big(C^{(s,t)};\ \Delta^{(s,t)}(F_R^{(s,t)})\big),
\end{equation}
where $\Delta^{(s,t)}(\cdot)$ denotes the learned convolutional offset generator. By adapting sampling locations to the underlying reference structures, the module aligns feature comparison to semantically meaningful regions. The receptive field can expand in smooth areas to accumulate broader evidence (where distortions are typically less salient), while focusing on local structural consistency around edges or fine textures. This behavior is consistent with content-dependent masking and attention effects in visual perception~\cite{chandler2013seven}, enabling CAFM to implicitly learn spatially varying perceptual weights.

\subsubsection{Multi-Scale Fusion}

Different scales capture complementary cues: shallow features preserve fine spatial details, whereas deep features encode more semantic and context-aware information. We employ an attention-based MSF module inspired by CBAM~\cite{woo2018cbam} to refine and integrate $\{H^{(s,t)}\}_{s=1}^{4}$.

\textbf{Attention refinement.} For each scale, we apply channel attention followed by spatial attention:

\begin{align}
H^{(s,t)}_{c}
&= H^{(s,t)} \odot \sigma\!\Big(
\text{MLP}(\text{AvgPool}(H^{(s,t)})) \notag\\
&\qquad + \text{MLP}(\text{MaxPool}(H^{(s,t)}))
\Big), \\
\tilde{H}^{(s,t)}
&= H^{(s,t)}_{c} \odot \sigma\!\Big(
\text{Conv}\big(
\text{Concat}\big(
\text{AvgPool}(H^{(s,t)}_{c}), \notag\\
&\qquad \text{MaxPool}(H^{(s,t)}_{c})
\big)
\big)
\Big),
\end{align}
where $\odot$ is element-wise multiplication and $\sigma(\cdot)$ is the sigmoid function. 
$\text{AvgPool}(\cdot)$ and $\text{MaxPool}(\cdot)$ denote global pooling operators.
$\text{MLP}(\cdot)$ is a multi-layer perceptron with one hidden layer.
Channel attention emphasizes distortion-relevant channels, while spatial attention highlights salient regions contributing to perceptual quality.

\textbf{Pooling and concatenation.} The refined features are compressed into scale-specific vectors via global average pooling:
\begin{equation}
v^{(s,t)}=\text{AvgPool}\big(\tilde{H}^{(s,t)}\big),
\end{equation}
and concatenated to form a unified multi-scale representation:
\begin{equation}
V^{(t)}=\text{Concat}\big(v^{(1,t)},v^{(2,t)},v^{(3,t)},v^{(4,t)}\big).
\end{equation}

\subsubsection{Quality Regression}

We regress the frame-level quality score using a three-layer MLP:
\begin{equation}
\hat{q}^{(t)}=\text{FC}_3\!\left(\text{ReLU}\!\left(\text{FC}_2\!\left(\text{ReLU}\!\left(\text{FC}_1\!\left(V^{(t)}\right)\right)\right)\right)\right).
\end{equation}
Video-level quality is obtained by temporal average pooling:
\begin{equation}
\hat{q}_{\text{Deep}}=\frac{1}{T}\sum_{t=1}^{T}\hat{q}^{(t)},
\end{equation}
where $T$ denotes the number of sampled frames. 

\subsection{Traditional Metric Branch}

Our empirical study with different hand-crafted metrics indicates that VMAF provides the strongest complement to the deep feature component.
VMAF integrates multiple elementary quality features through Support Vector Regression (SVR).

\begin{itemize}
\item \textbf{Detail Loss Metric (DLM):} Quantifies loss of detail and texture
\item \textbf{Visual Information Fidelity (VIF):} Measures information fidelity in the HVS framework  
\item \textbf{Temporal Information (TI):} Captures motion-related features based on frame differences
\end{itemize}

\subsection{Extension to HDR Video Quality Assessment}

For HDR video quality assessment, we apply Perceptual Uniform encoding (PU21)~\cite{pu21} as a preprocessing step before feeding videos into the deep feature component of FDIM.

For HDR input $(R_{\text{HDR}}, D_{\text{HDR}})$, we first convert display-encoded RGB values into absolute linear colorimetric intensities by accounting for the display peak luminance, black level, and ambient light reflection. Let $I_{\mathrm{de},c}(x)$ denote the encoded pixel value at location $x$ for channel $c\in\{R,G,B\}$, and let $E(\cdot)$ be the electro-optical transfer function (EOTF) of the signal format (e.g., sRGB for SDR or PQ/HLG for HDR). We compute
\begin{equation}
\begin{split}
I_{\mathrm{lin},c}(x)
&= \min\Big\{(L_{\mathrm{peak}}-L_{\mathrm{black}})\,E\big(I_{\mathrm{de},c}(x)\big)+L_{\mathrm{black}}, \notag\\
&\qquad L_{\mathrm{peak}}\Big\} + L_{\mathrm{refl}}
\end{split}
\label{eq:display_model}
\end{equation}
where $L_{\mathrm{peak}}$ and $L_{\mathrm{black}}$ are the display peak luminance and black level (cd/m$^2$). The reflected ambient component is
\begin{equation}
L_{\mathrm{refl}}=k_{\mathrm{refl}}\frac{E_{\mathrm{amb}}}{\pi},
\label{eq:ambient_reflection} 
\end{equation}
with screen reflectivity $k_{\mathrm{refl}}$ and ambient illumination $E_{\mathrm{amb}}$ in lux. For PQ signals, since the PQ EOTF maps code values directly to absolute luminance (cd/m$^2$), the multiplicative factor $(L_{\mathrm{peak}}-L_{\mathrm{black}})$ can be omitted, and values are clipped at $L_{\mathrm{peak}}$.
We then apply the PU21 transfer function independently to each channel:
\begin{equation}
I_{\mathrm{PU21},c}(x) = \text{PU21}(I_{\mathrm{lin},c}(x))
\end{equation}
The PU21 function is a piecewise polynomial that approximates a perceptually uniform mapping.
Finally, each channel of $I_{\mathrm{PU21},c}(x)$ is normalized to [0,1] for network input.

\subsection{Training Strategy}

\subsubsection{Training Data}
\label{sec:training_data}

We trained and validated our model on the Diverse Compressed Video Quality Assessment dataset (DCVQA), collaboratively constructed by multiple institutions under agreed license compliance protocols~\cite{TVQA}. The dataset contains 1{,}070 source contents in total: 1{,}000 source contents provided by one collaborator, 50 source contents from six open-source video datasets (e.g., CDVL~\cite{CDVL}, Xiph~\cite{Xiph}, IVP~\cite{IVP}, LIVE-NRLX-II~\cite{LIVE-NRLX-II}, and GamingVideoSET~\cite{GamingVideoSET}), and 20 source contents provided by the Academy of Broadcasting Science of the National Radio and Television Administration (NRTA).
For this study, 200 collaborator source contents were reserved as a held-out test set and were not used for training or validation. The training split used the remaining 800 collaborator source contents, while the 70 external source contents (50 open-source + 20 NRTA) were allocated for validation, with no scene or content overlap with the training set.
All video clips were standardized to a consistent acquisition and representation format: 1920$\times$1080 resolution, YUV 4:2:0 chroma sampling (yuv420p), and a 25 fps frame rate.

The dataset covers both conventional and neural video codecs.
For conventional coding, two representative standards are included: H.265/HEVC (open-source encoder \texttt{x265}) and AVS3 (reference encoder \texttt{uavs3e}).
Each source is encoded at 4 operating points per codec using CRF/QP configurations.
In addition, four end-to-end neural codecs are included.
EEM\footnote{EEM is an NVC exploration model under development by the Audio Video coding Standard (AVS) Workgroup of China. The reference software is available at: \url{https://gitlab.com/xhsheng/avs-eem}} and AlphaVC-P~\cite{shi2022alphavc,alphavc-p} are encoded at four target bitrates, while two private NVC codecs are evaluated at two and three operating points, respectively. 
EEM is an end-to-end neural codec optimized for MSE. 
AlphaVC-P is a perceptually optimized codec trained with a combination of perceptual losses (including LPIPS and GAN-based~\cite{goodfellow2014generativeadversarialnetworks} losses) and Mean Square Error (MSE) loss. 

Unlike traditional video codecs that primarily introduce degradations relative to the source, NVCs can occasionally yield perceptual quality enhancement. To capture both degradations and potential enhancements during subjective data collection, we adopted a Degradation Category Rating (DCR) method with a 7-point scale, which extends the 5-point impairment scale by adding two positive preference levels:
\{7: clearly better, 6: slightly better, 5: imperceptible, 4: perceptible but not annoying, 3: slightly annoying, 2: annoying, 1: very annoying\}.
The subjective database construction follows a hybrid protocol that combines crowdsourcing and controlled laboratory tests.
Specifically, 500 source videos with the highest bandwidth requirements were evaluated in a laboratory environment, while the remaining 570 source videos were assessed via online crowdsourcing.
The MOS collected under the two settings were statistically consistent.
All subjective scores were aggregated to form a large-scale dataset.
For model training, we excluded samples with MOS above 5 and focused on degradation prediction.

\subsubsection{Data Pairing Strategy}

Following established practices in learning-to-rank for quality assessment~\cite{ma2017dipiq}, we adopt a pairwise comparison strategy to improve the ability of the metric to differentiate the quality of different videos.

For each sample tuple $(R, D, \mu, \sigma)$ where $R$ is the reference, $D$ is the distorted video, $\mu$ is the MOS, and $\sigma$ is the score standard deviation, we generate two types of pairs:

\textbf{Homogeneous Pairs:} From sequences sharing the same reference $R_k$, we sample pairs:
\begin{equation}
(R_k, D_{k,i}, \mu_{k,i}, \sigma_{k,i}), \quad (R_k, D_{k,j}, \mu_{k,j}, \sigma_{k,j}), \quad i \neq j
\end{equation}

\textbf{Heterogeneous Pairs:} From sequences with different references $R_k, R_m$ ($k \neq m$), we sample:
\begin{equation}
(R_k, D_{k,i}, \mu_{k,i}, \sigma_{k,i}), \quad (R_m, D_{m,j}, \mu_{m,j}, \sigma_{m,j})
\end{equation}

Assuming video quality follows a Gaussian distribution with mean $\mu$ and variance $\sigma^2$, and that the qualities of different videos are independent, the quality difference between videos $D_{k,i}$ and $D_{m,j}$ follows:
\begin{equation}
\Delta Q \sim \mathcal{N}(\mu_{k,i} - \mu_{m,j}, \sigma_{k,i}^2 + \sigma_{m,j}^2)
\end{equation}

The ground-truth preference probability that $D_{k,i}$ is better than $D_{m,j}$ is computed via the Gaussian cumulative distribution:
\begin{equation}
g_{i,j} = \Phi\left(\frac{\mu_{k,i} - \mu_{m,j}}{\sqrt{\sigma_{k,i}^2 + \sigma_{m,j}^2}}\right)
\end{equation}
where $\Phi(\cdot)$ is the Cumulative Distribution Function (CDF) of the standard normal distribution.

\subsubsection{Loss Function}

We employ a pairwise fidelity (ranking) loss. For a pair, the model predicts a quality score $\hat{q}$ and an uncertainty $\hat{\sigma}$. The predicted preference probability is:
\begin{equation}
\hat{p}_{i,j} = \Phi\left(\frac{\hat{q}_i - \hat{q}_j}{\sqrt{\hat{\sigma}_i^2 + \hat{\sigma}_j^2}}\right)
\end{equation}
We then minimize the divergence between the predicted distribution $\hat{p}_{i,j}$ and the ground-truth distribution $g_{i,j}$ using a fidelity loss:
\begin{equation}
\mathcal{L}_{\text{rank}} = 1 - g_{i,j}\cdot \hat{p}_{i,j} - (1-g_{i,j})\cdot(1-\hat{p}_{i,j})
\end{equation}

\section{Experimental Setup}
\label{sec:experiments}

\subsection{Implementation Details}

\begin{table*}[t]
\centering
\caption{Summary of the datasets used in our experiments.}
\label{tab:datasets_summary}
\setlength{\tabcolsep}{3.2pt}
\renewcommand{\arraystretch}{1.15}
\resizebox{\textwidth}{!}{
\begin{tabular}{l|ccccc|ccccc}
\toprule
& \multicolumn{5}{c|}{\textbf{SDR}} & \multicolumn{5}{c}{\textbf{HDR}} \\
\cmidrule(lr){2-6}\cmidrule(lr){7-11}
\textbf{Attribute}
& \makecell{\textbf{DCVQA}}
& \makecell{\textbf{MCML}\\\textbf{4K}}
& \makecell{\textbf{AVT-VQDB}\\\textbf{UHD-1}}
& \makecell{\textbf{Waterloo}\\\textbf{IVC 4K}}
& \textbf{BVI-HD}
& \textbf{ZJUHDR}
& \makecell{\textbf{AVT-VQDB}\\\textbf{UHD-1-HDR}}
& \textbf{HDR-VDC}
& \textbf{LIVE-HDR}
& \textbf{HDRSDR-VQA}
\\
\midrule
Number (Ref)
& 1070 & 10 & 17 & 20 & 32 & \makecell{8} & 5 & 16 & 31 & 54 \\
Resolutions (Ref)
& \makecell{1920$\times$1080}
& \makecell{3840$\times$2160}
& \makecell{3840$\times$2160}
& \makecell{3840$\times$2160}
& \makecell{1920$\times$1080}
& \makecell{3840$\times$2160}
& \makecell{3840$\times$2160}
& \makecell{3840$\times$2160}
& \makecell{3840$\times$2160}
& \makecell{3840$\times$2160}
\\
Frame rates (Ref)
& 25 & 30 & 60 & \makecell{24/25/30} & \makecell{30/50/60}
& \makecell{60} & 60 & \makecell{24/25/30/60} & \makecell{50/60} & 60 \\
Duration (s)
& 8 & 10 & \makecell{8/10} & 10 & 5 & \makecell{10} & 10 & \makecell{5--11} & \makecell{7--10} & 7 \\
\midrule
Number (Dist)
& \makecell{16,136 for training\\1,299 for evaluation} & 240 & \makecell{756\\(218 evaluated)} & 1,200 & 384 & \makecell{178} & 200 & 132 & 310 & \makecell{480\\(180 evaluated)} \\
Resolutions (Dist)
& \makecell{1920$\times$1080}
& \makecell{1920$\times$1080\\3840$\times$2160}
& \makecell{360p\\480p\\720p\\1080p\\1440p\\2160p}
& \makecell{960$\times$540\\1920$\times$1080\\3840$\times$2160}
& \makecell{1920$\times$1080}
& \makecell{3840$\times$2160}
& \makecell{720p\\1080p\\1440p\\4K}
& \makecell{4K\\1080p\\720p}
& \makecell{4K\\1080p\\720p\\540p}
& \makecell{540p\\720p\\1080p\\1440p\\4K}
\\
Frame rates (Dist)
& 25 & 30 & \makecell{15/24/30/60} & \makecell{24/25/30} & \makecell{30/50/60}
& \makecell{60} & 60 & \makecell{24/25/30/60} & \makecell{50/60} & 60 \\
Distortion types
& \makecell{compression\\(conventional \& neural)}
& \makecell{compression\\spatial scaling}
& \makecell{compression\\spatial scaling\\temporal scaling}
& \makecell{compression\\spatial scaling}
& \makecell{compression\\synthesis}
& \makecell{compression\\(conventional \& neural)}
& \makecell{compression\\spatial scaling}
& \makecell{compression\\spatial scaling}
& \makecell{compression\\spatial scaling}
& \makecell{compression\\spatial scaling}
\\
Encoding standard
& \makecell{HEVC\\AVS3\\EEM\\AlphaVC-P\\Two private NVCs}
& \makecell{AVC\\HEVC\\VP9}
& \makecell{AVC\\HEVC\\VP9}
& \makecell{AVC\\HEVC\\VP9\\AV1\\AVS2}
& \makecell{HEVC}
& \makecell{VVC\\AVS3\\LCEVC\\EEM\\NNVC\\AlphaVC-P}
& \makecell{HEVC\\VVC\\AV1}
& \makecell{AV1}
& \makecell{HEVC}
& \makecell{HEVC}
\\
Encoder software
& \makecell{x265\\uavs3e\\EEM\\AlphaVC-P}
& \makecell{JM 18.5\\HM 10.0\\libvpx 1.3.0}
& \makecell{FFmpeg v3.2.2\\(libx264/libx265\\libvpx-vp9)}
& \makecell{x264, x265\\libvpx, aomenc\\libxavs2}
& \makecell{HM 14.0\\modified synthesis\\encoder}
& \makecell{VTM 23.9\\HPM 15.7\\LTM 7.0\\EEM 6.1\\NNVC\\AlphaVC-P}
& \makecell{FFmpeg v6.1\\(libx265/libaom-av1)\\VVenC} & \makecell{SVT-AV1} & \makecell{libx265} & \makecell{libx265}
\\
\makecell[l]{Distorted versions\\per source}
& \makecell{$\sim$20\\(6 encoders\\$\times$levels)}
& \makecell{24\\(3 encoders\\$\times$2 res.\\$\times$4 levels)}
& \makecell{24/30/64\\(3 encoders\\$\times$6 res.\\$\times$fps\\$\times$levels)}
& \makecell{60\\(5 encoders\\$\times$3 res.\\$\times$4 levels)}
& \makecell{12\\(2 encoders\\$\times$6 levels)}
& \makecell{$\sim$24\\(6 encoders\\$\times$ 4 levels)}
& \makecell{40\\(3 encoders\\$\times$4 res.\\$\times$bitrates)}
& \makecell{10\\(1 encoder\\$\times$3 res.\\$\times$3 bitrates)}
& \makecell{9\\(1 encoder\\$\times$4 res.\\$\times$bitrates)}
& \makecell{9\\(1 encoder\\$\times$4 res.\\$\times$bitrates)}
\\
\midrule
Subjective method
& \makecell{DCR}
& \makecell{ACR-HR}
& \makecell{ACR}
& \makecell{ACR}
& \makecell{DSCQS}
& \makecell{DCR} & \makecell{ACR-HR} & \makecell{PC} & \makecell{ACR-HR} & \makecell{PC}
\\
Rating scale
& \makecell{7-point\\discrete}
& \makecell{10-point\\discrete}
& \makecell{5-point\\discrete}
& \makecell{100-point\\continuous}
& \makecell{100-point\\continuous}
& \makecell{5-point\\discrete} & \makecell{5-point\\discrete} & \makecell{continuous\\(scaled to JOD)} & \makecell{100-point\\continuous} & \makecell{Continuous\\(scaled to JOD)}
\\
Viewing environment
& \makecell{crowdsourced\\+ controlled\\laboratory}
& \makecell{controlled\\laboratory}
& \makecell{controlled\\laboratory}
& \makecell{controlled\\laboratory}
& \makecell{controlled\\laboratory}
& \makecell{controlled\\laboratory} & \makecell{controlled\\laboratory} & \makecell{controlled\\laboratory} & \makecell{controlled\\laboratory} & \makecell{controlled\\laboratory}
\\
\bottomrule
\end{tabular}
}
\end{table*}

We trained the deep feature component of FDIM using Adam with $\beta_1 = 0.9$ and $\beta_2 = 0.999$, an initial learning rate of $1\times10^{-4}$, weight decay of $5\times10^{-4}$, and a batch size of 8 video pairs. 
During training, we uniformly sampled one frame per second from each video clip. Data augmentation included random spatial crops (512$\times$512) and horizontal flipping ($p=0.5$). 
Pairwise sampling substantially expanded the number of training instances, and each video participated in roughly 20 distinct pairs within one epoch. 
We implemented the deep feature component of FDIM in PyTorch and trained the model for one epoch.
During evaluation, videos were processed at their original resolution; when needed, distorted videos were upsampled to match the spatial resolution of their reference videos.
We fitted each component-specific mapping in Equation~\ref{eq:branch_mapping} using subjective scores from DCVQA and the corresponding component outputs, ensuring the mapping function accurately reflected human perceptual judgments. These parameters are fixed for inference.

\subsection{Evaluation Datasets}

We evaluated our method on the DCVQA validation set as well as on multiple public benchmarks spanning both SDR and HDR content and diverse codec types. All datasets were evaluated using the same FDIM model weights without any dataset-specific fine-tuning. The evaluation datasets are summarized in Table~\ref{tab:datasets_summary}.
Detailed descriptions of the evaluation datasets and dataset-specific preprocessing are provided in the supplementary material.

\subsection{Evaluation Metrics and Protocols}

We employed two standard correlation metrics to measure the agreement between objective predictions and subjective scores: Pearson Linear Correlation Coefficient (PLCC) and Spearman Rank-Order Correlation Coefficient (SROCC). Following~\cite{mapping_PLCC}, we fitted a five-parameter logistic function to map each method's predicted scores to subjective scores before computing PLCC. This evaluation mapping was applied uniformly to all compared methods and is independent of the component-specific mapping in Equation~\ref{eq:branch_mapping}. Higher values indicate stronger correlation with human perception.
Following standard practice in quality assessment research, we reported results under two evaluation settings: per-sequence evaluation, where sequences are grouped by reference content and performance is computed within each group and averaged, and all-sequence evaluation, where performance is computed across all sequences. Per-sequence results are reported in the supplementary material.

\subsection{Compared Methods}

We considered the following metrics for comparison in both SDR and HDR evaluation. For SDR, we included PSNR, XPSNR, SSIM~\cite{wang2004image} and MS-SSIM~\cite{wang2003multiscale} as pixel-based metrics; VMAF~\cite{li2016toward} as a fusion-based metric; LPIPS~\cite{zhang2018unreasonable}, DISTS~\cite{ding2020image}, TOPIQ~\cite{chen2024topiq} and RankDVQA~\cite{feng2024rankdvqa} as representative deep learning-based metrics. For HDR, we evaluated wPSNR, DeltaE100 and PSNRL100~\cite{Segall2023JVETAC2011} as HDR-specific pixel-based metrics; PSNR+PU21 and SSIM+PU21 as traditional metrics with PU21 preprocessing; HDR-VDP-3.0.7~\cite{hdrvdp32023arXiv}, HDRMAX+VMAF~\cite{LIVEHDR2024TIP} and ColorVideoVDP~\cite{colorvideovdp} as representative metrics for HDR. 
Implementation details of the compared methods are provided in the supplementary material.

\section{Results and Analysis}

\label{sec:results}

\subsection{Performance on NVC VQA Datasets (DCVQA and ZJUHDR)}

\begin{table}[t]
\centering
\caption{Correlation results on DCVQA under all-sequence evaluation.}
\label{tab:avs_val_set3_codecgroup}
\small
\begin{tabular}{lcccc}
\toprule
Method & \multicolumn{2}{c}{Neural} & \multicolumn{2}{c}{Traditional} \\
\cmidrule(lr){2-3} \cmidrule(lr){4-5}
& PLCC & SROCC & PLCC & SROCC \\
\midrule

PSNR & 0.4761 & 0.4680 & 0.5077 & 0.5429 \\
XPSNR & 0.7510 & 0.6882 & 0.7836 & 0.7710 \\
SSIM & 0.5427 & 0.4514 & 0.5048 & 0.4989 \\
MS-SSIM & 0.7219 & 0.6470 & 0.7293 & 0.7365 \\
VMAF & 0.8060 & 0.7304 & 0.8913 & 0.8565 \\
\midrule
LPIPS & 0.7453 & 0.6871 & 0.6920 & 0.6905 \\
DISTS & 0.7714 & 0.6995 & 0.7499 & 0.7323 \\
TOPIQ & 0.8514 & 0.7989 & 0.8626 & 0.8419 \\
RankDVQA & 0.8203 & 0.7457 & 0.8725 & 0.8494 \\
\midrule
FDIM(Deep) & \textbf{0.9012} & \textbf{0.8310} & \underline{0.9190} & \underline{0.8898} \\
FDIM & \underline{0.9010} & \underline{0.8211} & \textbf{0.9312} & \textbf{0.8978} \\
\bottomrule
\end{tabular}
\end{table}

Table~\ref{tab:avs_val_set3_codecgroup} reports correlations on the DCVQA validation set, split by codec group (Neural / Traditional).
The performance results on the full validation set are reported in Table~\ref{tab:dcvqa_sdr_allseq}.
FDIM and FDIM(Deep) demonstrate strong performance, consistently outperforming both widely-used traditional metrics (e.g., PSNR, SSIM, VMAF) and recent deep learning-based methods (e.g., LPIPS, DISTS, TOPIQ, RankDVQA) across evaluation protocols. 
Existing metrics exhibit varying degrees of performance degradation on neural codecs compared with traditional codecs, while FDIM and FDIM(Deep) maintain more consistent performance.
The performance gain of FDIM over the deep feature component is most pronounced on traditional codecs, underscoring the complementary role of VMAF in modeling conventional coding distortions. Conversely, the improvement of FDIM over VMAF is more significant on neural codecs, highlighting the effectiveness of the deep feature component in capturing complex generative artifacts.
These findings validate the effectiveness of the proposed content-adaptive feature-distance modeling mechanism and the hybrid fusion of hand-crafted and deep features, resulting in accurate quality prediction.

\begin{table}[t]
\centering
\caption{Correlation results on ZJUHDR under codec-group evaluation. The symbol * indicates that the predictions were computed at a temporal sampling rate of 0.25 frames per second for HDR-VDP-3.0.7.}
\label{tab:zjuhdr_phase1_codecgroup}
\small
\begin{tabular}{lcccc}
\toprule
Method & \multicolumn{2}{c}{Neural} & \multicolumn{2}{c}{Traditional} \\
\cmidrule(lr){2-3} \cmidrule(lr){4-5}
& PLCC & SROCC & PLCC & SROCC \\
\midrule
PSNR & 0.5867 & 0.6053 & 0.6328 & 0.6137 \\
SSIM & 0.3921 & 0.4281 & 0.4188 & 0.4277 \\
VMAF & 0.8144 & 0.8353 & \underline{0.9054} & 0.8897 \\
\midrule
LPIPS & 0.7712 & 0.7359 & 0.8142 & 0.8796 \\
DISTS & 0.7090 & 0.6823 & 0.8725 & 0.8945 \\
TOPIQ & 0.8196 & 0.8248 & 0.8588 & 0.8772 \\
RankDVQA & 0.8203 & 0.8340 & 0.8563 & 0.8585 \\
\midrule
wPSNR & 0.6272 & 0.6224 & 0.5186 & 0.5302 \\
DeltaE100 & 0.5321 & 0.5112 & 0.4621 & 0.4500 \\
PSNRL100 & 0.5162 & 0.5217 & 0.3970 & 0.3976 \\
PSNR+PU21 & 0.6562 & 0.6219 & 0.6656 & 0.6224 \\
SSIM+PU21 & 0.5618 & 0.6116 & 0.5667 & 0.5776 \\
HDR-VDP-3.0.7* & \underline{0.8859} & \underline{0.8866} & 0.9002 & 0.8748 \\
HDRMAX+VMAF & 0.8568 & 0.8229 & 0.8097 & 0.7913 \\
ColorVideoVDP & 0.7769 & 0.7855 & 0.8236 & 0.8250 \\
\midrule
FDIM(Deep)+PU21 & 0.8550 & 0.8462 & 0.9043 & \underline{0.8990} \\
FDIM+PU21 & \textbf{0.9077} & \textbf{0.9069} & \textbf{0.9143} & \textbf{0.9051} \\
\bottomrule
\end{tabular}
\end{table}

Table~\ref{tab:zjuhdr_phase1_codecgroup} presents the codec-group evaluation results on ZJUHDR, split by codec group (Neural / Traditional). ZJUHDR encompasses a diverse set of both traditional and learning-based codecs that are not included in the training data of FDIM. 
FDIM+PU21 consistently achieves the highest correlation with MOS across both subsets, outperforming not only conventional and deep learning-based SDR metrics but also existing HDR-oriented quality models. This result demonstrates strong capability in handling HDR-specific perceptual characteristics under diverse and previously unseen compression distortions. 
Furthermore, the deep branch of FDIM significantly outperforms VMAF on neural codecs, highlighting its strong capability in modeling the content-varying artifacts introduced by learning-based compression. The hybrid fusion strategy further improves performance, indicating strong complementarity between deep and hand-crafted features for HDR quality assessment.

\subsection{Performance on SDR VQA Datasets}
\label{sec:public_vqa}

\begin{table*}[t]
\centering
\caption{Correlation between objective metrics and MOS on DCVQA and four public SDR datasets (all-sequence evaluation; Average is computed over all five datasets excluding CVQAC).}
\label{tab:dcvqa_sdr_allseq}
\small
\begin{tabular}{lcccccccccccc}
\toprule
Method & \multicolumn{2}{c}{DCVQA} & \multicolumn{2}{c}{MCML 4K} & \multicolumn{2}{c}{AVT-VQDB-UHD-1} & \multicolumn{2}{c}{WaterlooIVC 4K} & \multicolumn{2}{c}{BVI-HD} & \multicolumn{2}{c}{Average} \\
\cmidrule(lr){2-3} \cmidrule(lr){4-5} \cmidrule(lr){6-7} \cmidrule(lr){8-9} \cmidrule(lr){10-11} \cmidrule(lr){12-13}
 & PLCC & SROCC & PLCC & SROCC & PLCC & SROCC & PLCC & SROCC & PLCC & SROCC & PLCC & SROCC \\
\midrule
PSNR & 0.4808 & 0.4917 & 0.8264 & 0.8495 & 0.8774 & 0.8513 & 0.4844 & 0.4706 & 0.5849 & 0.5725 & 0.6508 & 0.6471 \\
XPSNR & 0.7549 & 0.7119 & 0.9569 & 0.9324 & 0.8991 & 0.9067 & 0.7856 & 0.7807 & 0.7815 & 0.7574 & 0.8356 & 0.8178 \\
SSIM & 0.5214 & 0.4658 & 0.8781 & 0.8872 & 0.8619 & 0.8462 & 0.5119 & 0.4978 & 0.6909 & 0.6711 & 0.6928 & 0.6736 \\
MS-SSIM & 0.7201 & 0.6823 & 0.8696 & 0.8697 & 0.6672 & 0.6209 & 0.5242 & 0.5198 & 0.6906 & 0.6624 & 0.6943 & 0.6710 \\
VMAF & 0.8347 & 0.7766 & 0.9565 & 0.9161 & 0.9238 & 0.9120 & 0.7046 & 0.7113 & 0.7806 & 0.7724 & 0.8400 & 0.8177 \\
\midrule
LPIPS & 0.7255 & 0.6897 & 0.8605 & 0.8658 & 0.7506 & 0.7740 & 0.5317 & 0.5098 & 0.6792 & 0.6436 & 0.7095 & 0.6966 \\
DISTS & 0.7604 & 0.7129 & 0.8970 & 0.8869 & 0.8284 & 0.8285 & 0.7259 & 0.7134 & 0.7083 & 0.7123 & 0.7840 & 0.7708 \\
TOPIQ & 0.8524 & 0.8147 & 0.9562 & 0.9492 & 0.9414 & \textbf{0.9388} & 0.8000 & 0.7953 & 0.8139 & \underline{0.8000} & 0.8728 & 0.8596 \\
RankDVQA & 0.8491 & 0.7870 & 0.9186 & 0.9314 & 0.9305 & 0.9138 & 0.6619 & 0.6501 & 0.7539 & 0.7461 & 0.8228 & 0.8057 \\
\midrule
FDIM(Deep) & \underline{0.9070} & \textbf{0.8538} & \underline{0.9654} & \underline{0.9520} & \textbf{0.9580} & \underline{0.9365} & \textbf{0.8504} & \underline{0.8385} & \textbf{0.8229} & \textbf{0.8040} & \textbf{0.9007} & \underline{0.8770} \\
FDIM & \textbf{0.9135} & \underline{0.8529} & \textbf{0.9715} & \textbf{0.9612} & \underline{0.9504} & 0.9324 & \underline{0.8472} & \textbf{0.8398} & \underline{0.8191} & 0.7990 & \underline{0.9003} & \textbf{0.8771} \\
\bottomrule
\end{tabular}
\end{table*}

\begin{figure}[t]
\centering
\includegraphics[width=0.9\linewidth]{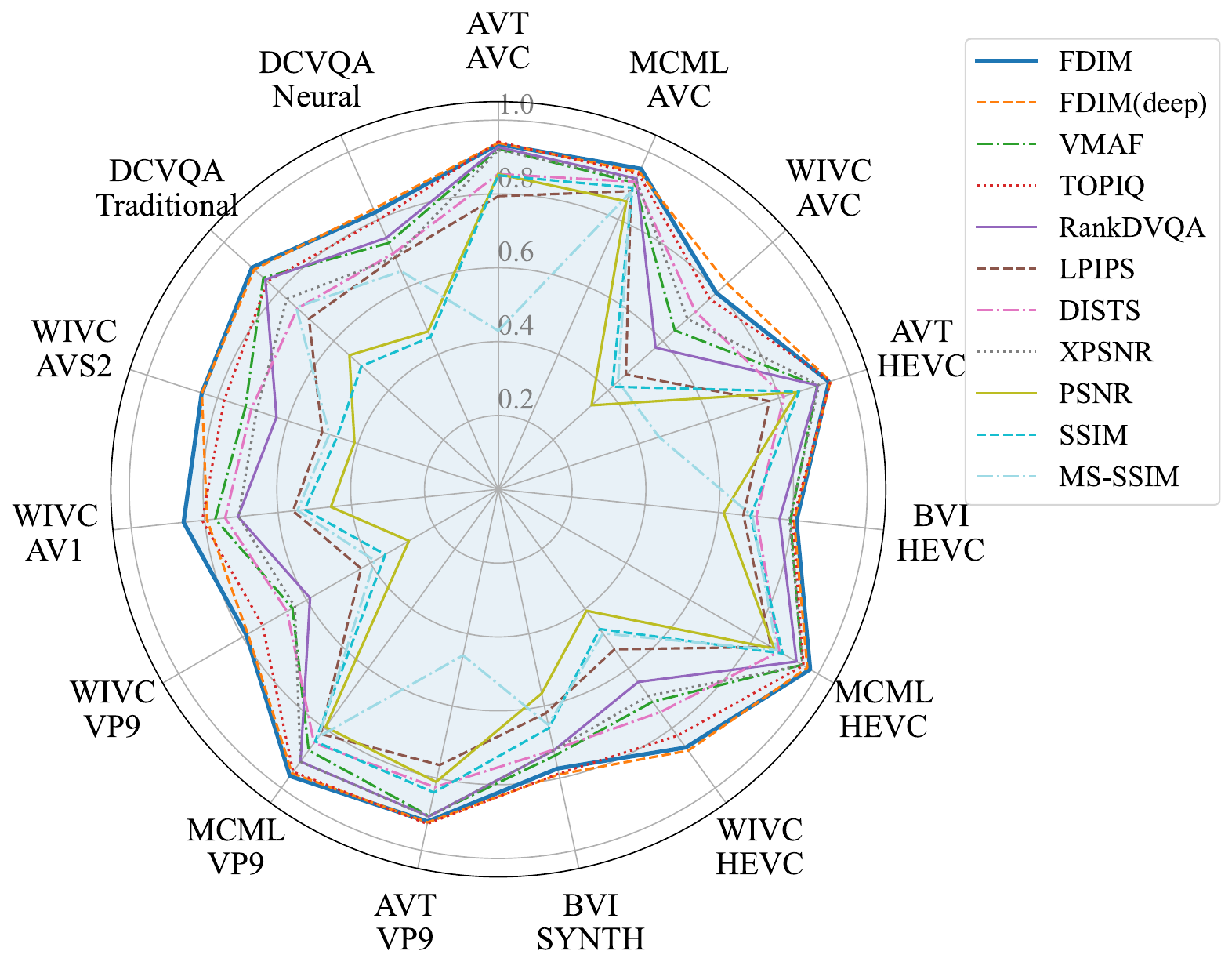}
\caption{SROCC performance comparison across different dataset-codec subsets. The axes represent specific combinations of datasets and codecs. MCML, AVT, WIVC, and BVI are abbreviations for MCML 4K, AVT-VQDB-UHD-1, WaterlooIVC 4K, and BVI-HD, respectively. FDIM consistently achieves high correlations across diverse compression methods.}
\label{fig:radar_all}
\end{figure}

We further evaluate FDIM on four public SDR datasets, including three 4K UHD benchmarks (MCML 4K, AVT-VQDB-UHD-1, and WaterlooIVC 4K) and one HD benchmark (BVI-HD). Table~\ref{tab:dcvqa_sdr_allseq} reports the correlations between objective scores and MOS, and also provides the average performance over five SDR datasets. 
The average performance gap between FDIM and FDIM(Deep) is relatively small, indicating that the deep feature component already captures distortions introduced by different conventional codecs, while the hand-crafted feature component further improves the overall generalization capability.
FDIM shows a significant performance advantage on the DCVQA and WaterlooIVC 4K datasets. Since these two datasets involve a wider range of codecs than the others, the results further support cross-codec generalization. FDIM also maintains performance advantages on the remaining three datasets.

To further analyze the robustness of FDIM across different coding methods and dataset characteristics, we visualize the SROCC performance of various methods in Fig.~\ref{fig:radar_all}. The axes represent different dataset-codec pairs (e.g., MCML-AVC, BVI-HEVC), and the radius indicates SROCC performance. As shown, FDIM (thick blue line) consistently forms the outermost envelope across most axes, indicating superior or highly competitive performance across almost all dataset-codec combinations. In contrast, competing methods such as LPIPS and RankDVQA exhibit greater performance instability, performing well on some subsets while degrading significantly on particular datasets or codec types. 
The overall performance of FDIM(Deep) and FDIM remains comparable and both significantly outperform other metrics, indicating that both the deep feature component and the hybrid framework can effectively generalize to diverse distortion types and content characteristics.

\subsection{Performance on HDR VQA Datasets}
\label{sec:hdr_results}

\begin{table*}[t]
\centering
\caption{Performance comparison of objective metrics on four public HDR datasets. The symbol * indicates that the predictions were computed at a temporal sampling rate of 1 frame per second, and 0.25 frames per second for HDR-VDP-3.0.7.}
\label{tab:hdr_overall_sp}
\scriptsize
\setlength{\tabcolsep}{3.5pt}
\begin{tabular}{lcccccccccccccccc}
\toprule
Method & \multicolumn{4}{c}{LIVE-HDR} & \multicolumn{2}{c}{\makecell{AVT-VQDB\\UHD-1-HDR}} & \multicolumn{8}{c}{HDR-VDC} & \multicolumn{2}{c}{HDRSDR-VQA} \\
\cmidrule(lr){2-5} \cmidrule(lr){6-7} \cmidrule(lr){8-15} \cmidrule(lr){16-17}
 & \multicolumn{2}{c}{bright} & \multicolumn{2}{c}{dark} & \multicolumn{2}{c}{-} & \multicolumn{2}{c}{bright+near} & \multicolumn{2}{c}{dim+near} & \multicolumn{2}{c}{bright+far} & \multicolumn{2}{c}{dim+far} & \multicolumn{2}{c}{-} \\
\cmidrule(lr){2-3} \cmidrule(lr){4-5} \cmidrule(lr){6-7} \cmidrule(lr){8-9} \cmidrule(lr){10-11} \cmidrule(lr){12-13} \cmidrule(lr){14-15} \cmidrule(lr){16-17}
 & PLCC & SROCC & PLCC & SROCC & PLCC & SROCC & PLCC & SROCC & PLCC & SROCC & PLCC & SROCC & PLCC & SROCC & PLCC & SROCC \\
\midrule
PSNR* & 0.7840 & 0.8140 & 0.7472 & 0.7708 & 0.6093 & 0.6181 & 0.6743 & 0.7059 & 0.6045 & 0.6617 & 0.5846 & 0.6812 & 0.5362 & 0.6284 & 0.7281 & 0.7554 \\
SSIM* & 0.6647 & 0.7870 & 0.6281 & 0.7525 & 0.4162 & 0.6206 & 0.6037 & 0.6744 & 0.5279 & 0.6189 & 0.5358 & 0.6329 & 0.4784 & 0.5723 & 0.4933 & 0.6981 \\
VMAF & 0.8962 & 0.9196 & 0.8763 & \underline{0.8955} & 0.8858 & \textbf{0.8927} & 0.8936 & 0.8945 & \underline{0.8699} & 0.8824 & \underline{0.8551} & 0.8517 & 0.8296 & 0.8226 & 0.8151 & 0.8324 \\
LPIPS* & 0.8339 & 0.8600 & 0.7965 & 0.8190 & 0.6679 & 0.6701 & 0.8043 & 0.8247 & 0.7639 & 0.7905 & 0.7530 & 0.7558 & 0.6973 & 0.6671 & 0.7461 & 0.7723 \\
DISTS* & 0.8497 & 0.8637 & 0.8372 & 0.8573 & 0.6940 & 0.7106 & 0.8203 & 0.8463 & 0.8124 & 0.8363 & 0.7728 & 0.7863 & 0.7414 & 0.7420 & 0.8188 & 0.8422 \\
TOPIQ* & 0.8837 & 0.9065 & 0.8429 & 0.8634 & 0.7365 & 0.7543 & \underline{0.9074} & \underline{0.9111} & 0.8612 & \underline{0.8992} & 0.8495 & \underline{0.8662} & \underline{0.8296} & 0.8256 & 0.8138 & 0.8336 \\
\midrule
wPSNR & 0.6747 & 0.6894 & 0.6188 & 0.6207 & 0.5449 & 0.5852 & 0.4717 & 0.5016 & 0.4505 & 0.4685 & 0.4523 & 0.5261 & 0.4368 & 0.5020 & 0.6924 & 0.7666 \\
DeltaE100 & 0.6876 & 0.6978 & 0.6488 & 0.6477 & 0.5859 & 0.5879 & 0.4462 & 0.4594 & 0.4174 & 0.4345 & 0.4332 & 0.4920 & 0.4148 & 0.4842 & 0.6691 & 0.6975 \\
PSNRL100 & 0.7243 & 0.7291 & 0.6938 & 0.6778 & 0.6935 & 0.6557 & 0.4840 & 0.4975 & 0.5053 & 0.4824 & 0.4988 & 0.5593 & 0.5130 & 0.5183 & 0.3160 & 0.0646 \\
PSNR+PU21* & 0.7858 & 0.8156 & 0.7503 & 0.7750 & 0.6744 & 0.6455 & 0.6471 & 0.6683 & 0.5859 & 0.6325 & 0.5572 & 0.6506 & 0.5224 & 0.6149 & 0.7406 & 0.7697 \\
SSIM+PU21* & 0.6898 & 0.7896 & 0.6602 & 0.7575 & 0.5268 & 0.6442 & 0.5801 & 0.6617 & 0.5172 & 0.6161 & 0.5215 & 0.6294 & 0.4770 & 0.5863 & 0.4922 & 0.7146 \\
HDR-VDP-3.0.7* & 0.7886 & 0.7979 & 0.7316 & 0.7360 & 0.6475 & 0.6534 & 0.8182 & 0.8331 & 0.7847 & 0.8121 & 0.7863 & 0.8105 & 0.7520 & 0.7513 & 0.6948 & 0.7133 \\
HDRMAX+VMAF* & 0.8703 & 0.8138 & 0.8783 & 0.8155 & \textbf{0.8944} & 0.8577 & 0.7464 & 0.6806 & 0.7565 & 0.6837 & 0.7301 & 0.6405 & 0.6978 & 0.6117 & 0.7321 & 0.7145 \\
ColorVideoVDP & 0.8821 & 0.8943 & 0.8383 & 0.8440 & 0.8573 & 0.8365 & 0.8425 & 0.8691 & 0.8138 & 0.8512 & 0.8509 & 0.8576 & 0.8136 & \underline{0.8269} & 0.7579 & 0.7808 \\
\midrule
FDIM(Deep)+PU21* & \textbf{0.9196} & \underline{0.9303} & \underline{0.8797} & 0.8923 & 0.8806 & 0.8515 & 0.8545 & 0.8887 & 0.8450 & 0.8848 & 0.8140 & 0.8465 & 0.7912 & 0.8185 & \textbf{0.8789} & \textbf{0.8950} \\
FDIM+PU21* & \underline{0.9177} & \textbf{0.9332} & \textbf{0.8847} & \textbf{0.8995} & \underline{0.8900} & \underline{0.8783} & \textbf{0.9204} & \textbf{0.9247} & \textbf{0.9085} & \textbf{0.9203} & \textbf{0.8917} & \textbf{0.8818} & \textbf{0.8788} & \textbf{0.8591} & \underline{0.8579} & \underline{0.8749} \\
\bottomrule
\end{tabular}
\end{table*}

We evaluated FDIM+PU21 on four public HDR VQA datasets, including LIVE-HDR, AVT-VQDB-UHD-1-HDR, HDR-VDC, and HDRSDR-VQA. Table~\ref{tab:hdr_overall_sp} reports the correlations (SROCC/PLCC). 
Overall, FDIM+PU21 achieves competitive correlations in most settings, indicating that the SDR-trained model can transfer to HDR content when combined with PU21 preprocessing. 
In contrast, many competing methods show large accuracy variations across different viewing conditions on HDR-VDC and LIVE-HDR, indicating weaker robustness to changes in viewing distance and ambient lighting. 
Moreover, FDIM+PU21 shows significant improvements over FDIM(Deep)+PU21 across multiple datasets and conditions, indicating that hybrid fusion improves transfer robustness.
Overall, these results indicate that the SDR-trained model can transfer to HDR content, yielding accurate and stable quality prediction across different HDR benchmarks.

\subsection{Ablation Studies}

\subsubsection{Effect of the scale of training data}
\label{sec:scale_ablation}

\begin{figure}[t]
  \centering
  \includegraphics[width=0.9\linewidth]{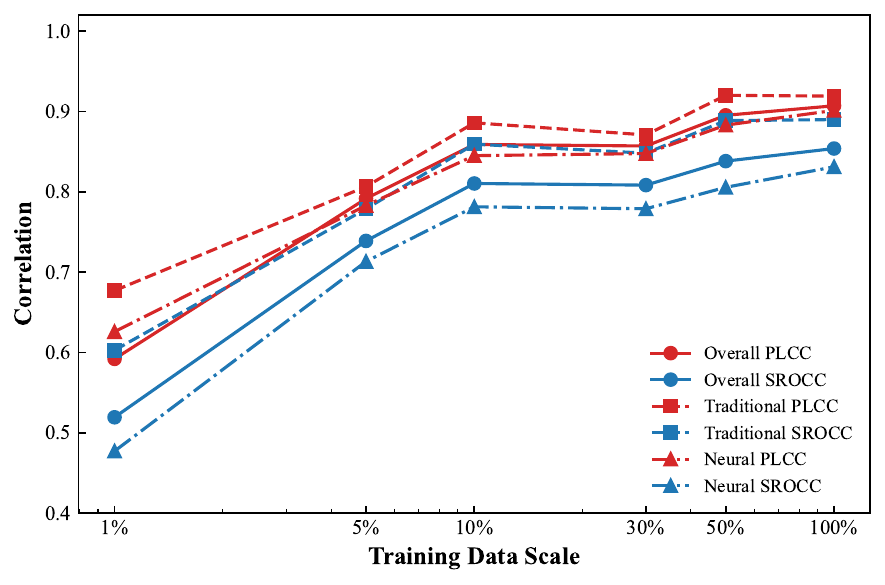}
  \caption{Performance of FDIM(Deep) with different scales of training data (1\%, 5\%, 10\%, 30\%, 50\%, 100\%). The results are reported in terms of PLCC and SROCC on the overall dataset, the traditional codec subset, and the neural codec subset.}
  \label{fig:ablation_scale}
\end{figure}

\begin{table*}[t]
\centering
\caption{Ablation experiment results on DCVQA dataset (all-sequence evaluation).}
\label{tab:ablation}
\setlength{\tabcolsep}{3pt}
\begin{tabular}{@{}llcccccc@{}}
\toprule
\multirow{2}{*}{Type} & \multirow{2}{*}{Configuration} & \multicolumn{2}{c}{Overall} & \multicolumn{2}{c}{Traditional} & \multicolumn{2}{c}{Neural} \\
\cmidrule(lr){3-4} \cmidrule(lr){5-6} \cmidrule(lr){7-8}
 &  & PLCC & SROCC & PLCC & SROCC & PLCC & SROCC \\
\midrule
\multicolumn{2}{c}{Default} & \textbf{0.9070} & \textbf{0.8538} & 0.9190 & \textbf{0.8898} & \textbf{0.9012} & \textbf{0.8310} \\
\midrule
\multirow{2}{*}{Training Data} & Neural \& Traditional codecs & 0.9004 ({\scriptsize -0.0066}) & 0.8385 ({\scriptsize -0.0153}) & \textbf{0.9225} ({\scriptsize 0.0035}) & 0.8875 ({\scriptsize -0.0023}) & 0.8866 ({\scriptsize -0.0146}) & 0.8064 ({\scriptsize -0.0246}) \\
 & Traditional codecs only & 0.8201 ({\scriptsize -0.0869}) & 0.7669 ({\scriptsize -0.0869}) & 0.9029 ({\scriptsize -0.0161}) & 0.8717 ({\scriptsize -0.0181}) & 0.8309 ({\scriptsize -0.0703}) & 0.7450 ({\scriptsize -0.0860}) \\
\midrule
\multirow{3}{*}{Offset} & Offset from distorted & 0.8547 ({\scriptsize -0.0523}) & 0.7953 ({\scriptsize -0.0585}) & 0.8656 ({\scriptsize -0.0534}) & 0.8401 ({\scriptsize -0.0497}) & 0.8580 ({\scriptsize -0.0432}) & 0.7744 ({\scriptsize -0.0566}) \\
 & Offset from discrepancy & 0.8858 ({\scriptsize -0.0212}) & 0.8337 ({\scriptsize -0.0201}) & 0.8946 ({\scriptsize -0.0244}) & 0.8682 ({\scriptsize -0.0216}) & 0.8854 ({\scriptsize -0.0158}) & 0.8143 ({\scriptsize -0.0167}) \\
 & Offset from concatenated & 0.8990 ({\scriptsize -0.0080}) & 0.8394 ({\scriptsize -0.0144}) & 0.9055 ({\scriptsize -0.0135}) & 0.8772 ({\scriptsize -0.0126}) & 0.8988 ({\scriptsize -0.0024}) & 0.8168 ({\scriptsize -0.0142}) \\
\midrule
\multirow{3}{*}{Component} & w/o discrepancy map & 0.8870 ({\scriptsize -0.0200}) & 0.8254 ({\scriptsize -0.0284}) & 0.9088 ({\scriptsize -0.0102}) & 0.8777 ({\scriptsize -0.0121}) & 0.8758 ({\scriptsize -0.0254}) & 0.7942 ({\scriptsize -0.0368}) \\
 & w/o deformable conv & 0.8878 ({\scriptsize -0.0192}) & 0.8329 ({\scriptsize -0.0209}) & 0.9038 ({\scriptsize -0.0152}) & 0.8737 ({\scriptsize -0.0161}) & 0.8831 ({\scriptsize -0.0181}) & 0.8114 ({\scriptsize -0.0196}) \\
 & w/o attention & 0.8721 ({\scriptsize -0.0349}) & 0.8265 ({\scriptsize -0.0273}) & 0.8688 ({\scriptsize -0.0502}) & 0.8513 ({\scriptsize -0.0385}) & 0.8777 ({\scriptsize -0.0235}) & 0.8115 ({\scriptsize -0.0195}) \\
\bottomrule
\end{tabular}
\end{table*}

To examine the data efficiency and scalability of the deep feature component, we evaluated its performance under varying proportions of the source content, specifically using 1\%, 5\%, 10\%, 30\%, 50\%, and 100\% of the 800 unique reference videos.
For each configuration, the training set consisted of the randomly selected reference videos and their associated distorted sequences. Accordingly, the pairwise sampling strategy was restricted to generating training pairs exclusively from these selected reference contents.
The results are illustrated in Fig.~\ref{fig:ablation_scale}.
We observe that the performance of the deep feature component improves with increasing training data, reflecting the scalable and data-driven nature of the model.
This trend demonstrates that the deep feature component of FDIM effectively leverages the diversity of video content to learn generalized quality representations.
Similar increasing trends are observed for both traditional and neural distortions, validating the effectiveness of the proposed deep features in capturing diverse distortion patterns when sufficient data is provided.

\begin{figure*}[t]
\centering
\includegraphics[width=\textwidth]{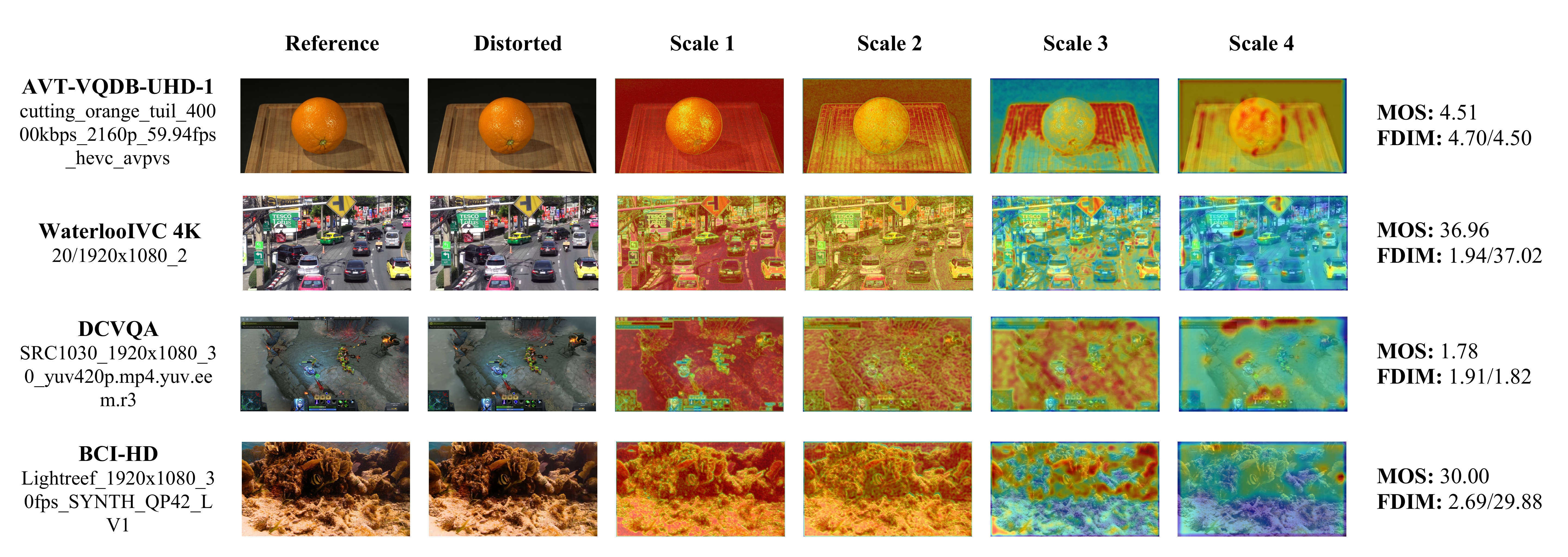}
\caption{Visualization of multi-scale features $\tilde{H}^{(s,t)}$. For each scale, feature maps are averaged along the channel dimension, normalized to $[0,1]$, and upsampled to the input resolution for overlay visualization. We report two scores for FDIM: the raw output score and a linearly mapped score using the dataset MOS scale, which facilitates observing the deviation from MOS.}
\label{fig:feature}
\end{figure*}

\subsection{Visualization of Multi-scale Features}
\label{sec:feature_vis}

\begin{figure}[t]
\centering
\includegraphics[width=0.9\linewidth]{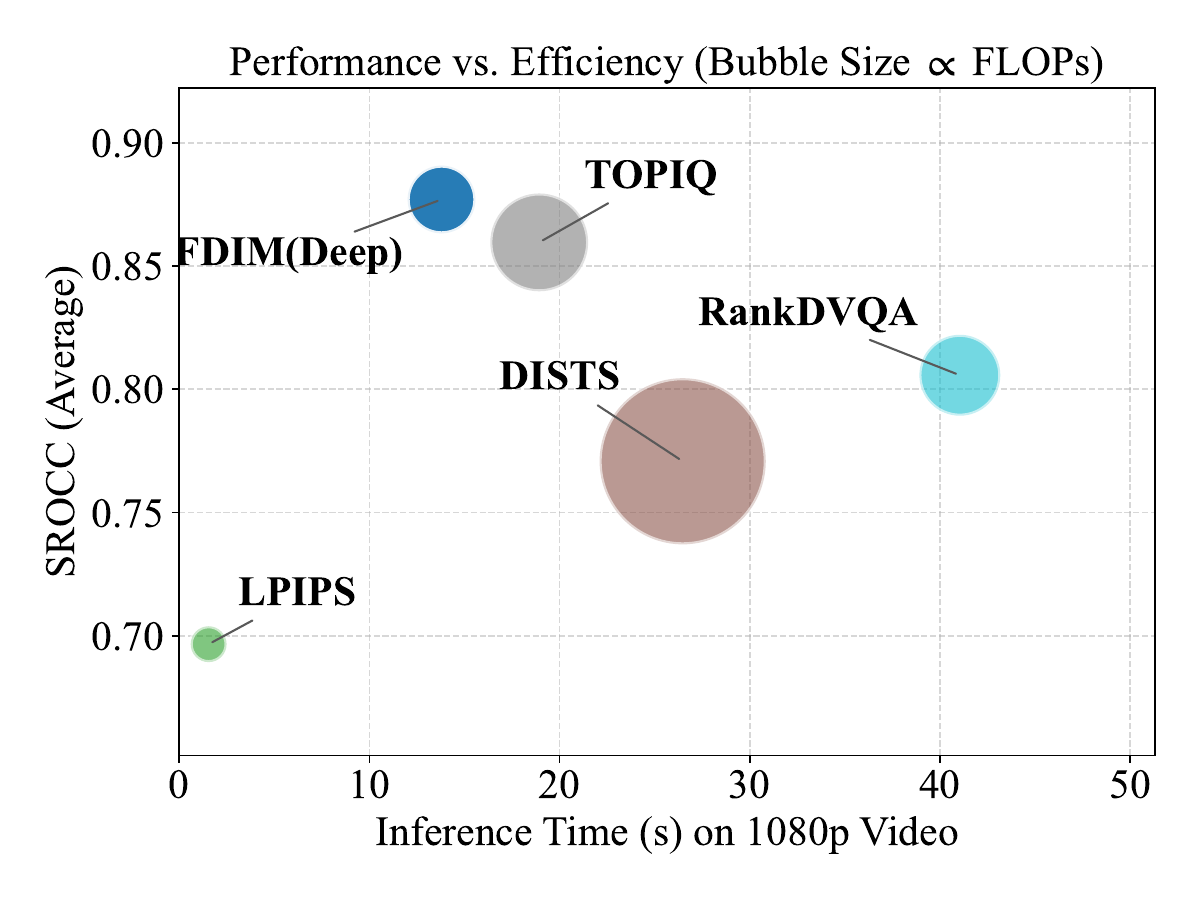}
\caption{Comparison of model performance (average SROCC on five SDR datasets) versus inference time and model size (bubble area). FDIM achieves an optimal balance between accuracy and efficiency.}
\label{fig:bubble_complexity}
\end{figure}

\begin{table*}[t]
\centering
\caption{Comparison of FLOPs and runtime (GPU; averaged over 100 runs) for a 150-frame video at different spatial resolutions.}
\label{tab:complexity}
\small
\begin{tabular}{lccccccc}
\toprule
Method & Params (M) & \multicolumn{2}{c}{720p} & \multicolumn{2}{c}{1080p} & \multicolumn{2}{c}{2160p} \\
\cmidrule(lr){3-4} \cmidrule(lr){5-6} \cmidrule(lr){7-8}
 &  & FLOPs (G) & Time (s) & FLOPs (G) & Time (s) & FLOPs (G) & Time (s) \\
\midrule
LPIPS & 2.47 & 26.31 & 0.96 & 59.48 & 1.55 & 241.06 & 6.57 \\
DISTS & 14.71 & 565.25 & 11.35 & 1272.66 & 26.47 & 5087.22 & 138.26 \\
TOPIQ & 33.27 & 185.95 & 8.83 & 441.40 & 18.93 & 2277.33 & 101.19 \\
RankDVQA & 4.59 & 86.29 & 11.42 & 302.00 & 41.04 & 1208.00 & 153.54 \\
\textbf{FDIM(Deep)} & 12.00 & 94.20 & 6.21 & 211.59 & 13.79 & 844.91 & 59.79 \\
\bottomrule
\end{tabular}
\end{table*}

As shown in Fig.~\ref{fig:feature}, we visualize the multi-scale features to interpret the representations used for quality prediction. 
In the deep feature component of FDIM, CAFM performs content-adaptive feature-distance modeling and produces discrepancy-aware representations.
MSF then applies channel and spatial gating to these representations, yielding $\tilde{H}^{(s,t)}$ that emphasizes perceptually informative responses across scales. The visualization reveals that shallow scales tend to exhibit localized activations around edges, textures, and subtle artifact patterns, while deep scales (with larger receptive fields) respond to broader regions associated with structural and global quality degradations. 

\subsubsection{Effect of the diversity of training data}
We further examined whether codec diversity in the training set is necessary for robust generalization.
Specifically, we retrained FDIM(Deep) using (1) a mixed set containing both neural and traditional codecs, and (2) a set containing only traditional codecs. 
The amount of training data was the same under both settings.
As reported in Table~\ref{tab:ablation}, the mixed setting preserves performance close to that of the default training setup, whereas using only traditional codecs leads to a pronounced degradation, with the largest drop observed on the neural subset.
These results highlight the importance of training on diverse codec types to learn transferable perceptual representations.

\subsubsection{Effect of Offset Learning Source}
To investigate the effectiveness of learning deformable offsets from the reference features $F_R^{(s,t)}$, we compared our proposed strategy with three alternatives: learning offsets from (1) distorted features $F_D^{(s,t)}$, (2) the discrepancy features $E^{(s,t)}$, and (3) the concatenated features $C^{(s,t)}$ of reference, distorted and discrepancy features.
The results in Table~\ref{tab:ablation} demonstrate that learning offsets from reference features yields the best performance.
Directly predicting offsets from distorted features results in the most significant performance drop, with PLCC and SROCC decreasing by 0.0523 and 0.0585, respectively. This suggests that distorted features, being corrupted by artifacts, may provide unreliable cues for aligning comparison features with semantically meaningful regions.
Using residual or concatenated features also leads to suboptimal results.
These findings validate that the pristine reference features provide the most reliable structural cues for guiding the deformable alignment, enabling more accurate discrepancy measurement.

\subsubsection{Impact of Network Components}
We further conducted ablation studies to verify the contribution of key components in FDIM, including the Content-Adaptive Feature-Distance Modeling (CAFM) module and the Multi-Scale Fusion (MSF) module.
\begin{itemize}
    \item \textbf{Discrepancy map in CAFM:} Removing the explicit discrepancy map calculation leads to a performance decline, with PLCC and SROCC dropping by 0.02 and 0.0284 on the overall dataset. This confirms that explicitly modeling feature differences facilitates the learning of distortion magnitude.
    \item \textbf{Deformable convolution in CAFM:} Replacing deformable convolution with standard convolution results in a PLCC/SROCC drop of 0.0192/0.0209. This highlights the importance of the deformable alignment in handling geometric mismatches and focusing on relevant spatial regions for quality assessment.
    \item \textbf{Attention mechanisms in MSF:} The removal of the attention mechanisms in the MSF module causes a notable degradation, particularly on traditional codecs where PLCC drops by 0.0502. The overall PLCC/SROCC decreases by 0.0349/0.0273, indicating that channel and spatial attention mechanisms are crucial for emphasizing perceptually significant features and suppressing irrelevant information.
\end{itemize}

\subsection{Complexity analysis}
To ensure a fair comparison of computational complexity, all tests were conducted on the same workstation. We evaluated the computation time (in seconds) for each method using 1080p videos with 150 frames.
The floating-point operations (FLOPs), parameter count, and inference time of the proposed method and other deep learning-based methods are summarized in Table~\ref{tab:complexity}. Fig.~\ref{fig:bubble_complexity} illustrates the trade-off between performance (SROCC), inference speed, and model size (represented by the bubble area). The reported SROCC values are consistent with the average results in Table~\ref{tab:dcvqa_sdr_allseq}.
Although RankDVQA has relatively low computational complexity, it exhibits lower accuracy and slower inference compared to our proposed deep feature component. TOPIQ, on the other hand, achieves higher accuracy but at a substantially higher computational cost. Overall, the proposed model achieves a favorable trade-off between accuracy and computational efficiency, indicating suitability for practical deployment.

\section{Conclusion}
\label{sec:conclusion}

In this work, we introduce FDIM, a generic full-reference video quality assessment method that integrates data-driven deep representations with hand-crafted features. The deep feature component employs content-adaptive feature-distance modeling and attention-based multi-scale fusion to capture distortions ranging from low-level fidelity degradation to content-varying neural artifacts, while the hand-crafted branch provides stable complementary cues grounded in conventional quality priors. This hybrid design yields robust quality prediction across diverse codecs, content types, and dynamic ranges.

Extensive evaluations on the DCVQA validation set and multiple public SDR and HDR benchmarks demonstrate that FDIM achieves strong performance and generalization under both all-sequence and per-sequence protocols. Across datasets, the deep branch provides strong transferability to diverse artifacts, and the hybrid fusion improves stability by leveraging complementary hand-crafted cues.
With a simple PU21 preprocessing, FDIM also demonstrates competitive SDR-to-HDR zero-shot transfer, indicating that the learned feature comparison mechanism captures perceptual discrepancies that generalize across dynamic ranges.
The ablation studies confirm the contribution of key components (CAFM and MSF) as well as the importance of training data scale and codec diversity for learning transferable perceptual representations. Our complexity analysis further shows a favorable trade-off between performance and computational efficiency, supporting practical deployment for video quality assessment. To facilitate more rigorous benchmarking of metric generalization, we will release the DCVQA validation set. Future work will focus on enhanced temporal modeling and further reducing computational cost to improve robustness and efficiency.

\bibliographystyle{IEEEtran}
\bibliography{ref}

@ARTICLE{Deep_Hierarchies_TPAMI,
  author={Kruger, Norbert and Janssen, Peter and Kalkan, Sinan and Lappe, Markus and Leonardis, Ales and Piater, Justus and Rodriguez-Sanchez, Antonio J. and Wiskott, Laurenz},
  journal={IEEE Transactions on Pattern Analysis and Machine Intelligence}, 
  title={Deep Hierarchies in the Primate Visual Cortex: What Can We Learn for Computer Vision?}, 
  year={2013},
  volume={35},
  number={8},
  pages={1847-1871},
  doi={10.1109/TPAMI.2012.272}}

@article{sullivan2012overview,
  author  = {Sullivan, Gary J. and Ohm, Jens-Rainer and Han, Woojin and Wiegand, Thomas},
  title   = {Overview of the High Efficiency Video Coding {(HEVC)} Standard},
  journal = {IEEE Transactions on Circuits and Systems for Video Technology},
  volume  = {22},
  number  = {12},
  pages   = {1649--1668},
  year    = {2012},
  doi     = {10.1109/TCSVT.2012.2221191}
}

@article{AV1,
  author   = {Han, Jingning and Li, Bohan and Mukherjee, Debargha and Chiang, Ching-Han and Grange, Adrian and Chen, Cheng and Su, Hui and Parker, Sarah and Deng, Sai and Joshi, Urvang and Chen, Yue and Wang, Yunqing and Wilkins, Paul and Xu, Yaowu and Bankoski, James},
  journal  = {Proceedings of the IEEE},
  title    = {A Technical Overview of {AV1}},
  year     = {2021},
  volume   = {109},
  number   = {9},
  pages    = {1435--1462},
  doi      = {10.1109/JPROC.2021.3058584}
}

@article{VVC,
  author       = {Benjamin Bross and
                  Ye{-}Kui Wang and
                  Yan Ye and
                  Shan Liu and
                  Jianle Chen and
                  Gary J. Sullivan and
                  Jens{-}Rainer Ohm},
  title        = {Overview of the Versatile Video Coding {(VVC)} Standard and its Applications},
  journal      = {{IEEE} Trans. Circuits Syst. Video Technol.},
  volume       = {31},
  number       = {10},
  pages        = {3736--3764},
  year         = {2021},
  doi          = {10.1109/TCSVT.2021.3101953},
  bibsource    = {dblp computer science bibliography, https://dblp.org}
}

@article{bross2021overview,
  author  = {Bross, Benjamin and Wang, Ye-Kui and Ye, Yan and Liu, Shan and Chen, Jianle and Sullivan, Gary J. and Ohm, Jens-Rainer},
  title   = {Overview of the Versatile Video Coding {(VVC)} Standard and Its Applications},
  journal = {IEEE Transactions on Circuits and Systems for Video Technology},
  volume  = {31},
  number  = {10},
  pages   = {3736--3764},
  year    = {2021},
  doi     = {10.1109/TCSVT.2021.3101953}
}

@inproceedings{shi2022alphavc,
  title     = {{AlphaVC}: High-Performance and Efficient Learned Video Compression},
  author    = {Shi, Yibo and Ge, Yunying and Wang, Jing and Mao, Jue},
  booktitle = {European Conference on Computer Vision},
  pages     = {616--631},
  year      = {2022},
  organization = {Springer}
}

@conference{alphavc-p,
    title = {Update on a perceptually optimized learning-based video codec for CVQM},
    organization = {ISO/IEC JTC 1/SC 29/AG 5 m69717},
    year = {2024},
}

@misc{goodfellow2014generativeadversarialnetworks,
  title         = {Generative Adversarial Networks},
  author        = {Goodfellow, Ian J. and Pouget-Abadie, Jean and Mirza, Mehdi and Xu, Bing and Warde-Farley, David and Ozair, Sherjil and Courville, Aaron and Bengio, Yoshua},
  year          = {2014},
  eprint        = {1406.2661},
  archivePrefix = {arXiv},
  primaryClass  = {stat.ML}
}

@inproceedings{lu2019dvc,
  author    = {Lu, Guo and Ouyang, Wanli and Xu, Dong and Zhang, Xiaoyun and Cai, Chunlei and Gao, Zhiyong},
  title     = {{DVC}: An End-to-End Deep Video Compression Framework},
  booktitle = {Proceedings of the IEEE/CVF Conference on Computer Vision and Pattern Recognition},
  pages     = {11006--11015},
  year      = {2019}
}

@inproceedings{hu2021fvc,
  author    = {Hu, Zhihao and Xu, Wei and Hu, Jiahao and Liu, Shuai and Yang, Wenhan and Lin, Weiyao},
  title     = {{FVC}: A New Framework towards Deep Video Compression in Feature Space},
  booktitle = {Proceedings of the IEEE/CVF Conference on Computer Vision and Pattern Recognition},
  pages     = {1502--1511},
  year      = {2021}
}

@InProceedings{Jia_2025_CVPR,
    author    = {Jia, Zhaoyang and Li, Bin and Li, Jiahao and Xie, Wenxuan and Qi, Linfeng and Li, Houqiang and Lu, Yan},
    title     = {Towards Practical Real-Time Neural Video Compression},
    booktitle = {Proceedings of the IEEE/CVF Conference on Computer Vision and Pattern Recognition (CVPR)},
    month     = {June},
    year      = {2025},
    pages     = {12543-12552}
}

@INPROCEEDINGS{DCVC_FM,
  author={Li, Jiahao and Li, Bin and Lu, Yan},
  booktitle={2024 IEEE/CVF Conference on Computer Vision and Pattern Recognition (CVPR)}, 
  title={Neural Video Compression with Feature Modulation}, 
  year={2024},
  volume={},
  number={},
  pages={26099-26108},
  doi={10.1109/CVPR52733.2024.02466}}

@misc{NNVC,
    title = {Algorithm Description for Neural Network-based Video Coding (NNVC-6.0)},
    organization = {Joint Video Experts Team (JVET)
of ITU-T SG16 WP3 and ISO/IEC JTC1/SC 29 AE2019-v2},
    year = {2023},
}

@article{ma2023overview,
  title={Overview of intelligent video coding: from model-based to learning-based approaches},
  author={Ma, Siwei and Gao, Junlong and Wang, Ruofan and Chang, Jianhui and Mao, Qi and Huang, Zhimeng and Jia, Chuanmin},
  journal={Visual Intelligence},
  volume={1},
  number={1},
  pages={15},
  year={2023},
  publisher={Springer}
}

@ARTICLE{JPEG_AI2024,
  author={Alshina, Elena and Ascenso, João and Ebrahimi, Touradj},
  journal={IEEE MultiMedia}, 
  title={JPEG AI: The First International Standard for Image Coding Based on an End-to-End Learning-Based Approach}, 
  year={2024},
  volume={31},
  number={4},
  pages={60-69},
  doi={10.1109/MMUL.2024.3485255}}

@article{wang2004image,
  author  = {Wang, Zhou and Bovik, Alan C. and Sheikh, Hamid R. and Simoncelli, Eero P.},
  title   = {Image Quality Assessment: From Error Visibility to Structural Similarity},
  journal = {IEEE Transactions on Image Processing},
  volume  = {13},
  number  = {4},
  pages   = {600--612},
  year    = {2004}
}

@article{sheikh2006image,
  author  = {Sheikh, Hamid R. and Bovik, Alan C.},
  title   = {Image Information and Visual Quality},
  journal = {IEEE Transactions on Image Processing},
  volume  = {15},
  number  = {2},
  pages   = {430--444},
  year    = {2006}
}

@misc{li2016toward,
  author       = {Li, Zhi and Aaron, Anne and Katsavounidis, Ioannis and Moorthy, Anush and Manohara, Megha},
  title        = {Toward a Practical Perceptual Video Quality Metric},
  howpublished = {Netflix Technology Blog},
  year         = {2016},
}

@ARTICLE{color_vmaf,
  author={Chen, L.-H. and Bampis, C. G. and Li, Z. and Sole, J. and Bovik, A. C.},
  journal={IEEE Transactions on Image Processing}, 
  title={Perceptual Video Quality Prediction Emphasizing Chroma Distortions}, 
  year={2021},
  volume={30},
  number={},
  pages={1408-1422},
  doi={10.1109/TIP.2020.3043127}}

@ARTICLE{ensemble_vmaf,
  author={Bampis, C. G. and Li, Z. and Bovik, A. C.},
  journal={IEEE Transactions on Circuits and Systems for Video Technology}, 
  title={Spatiotemporal Feature Integration and Model Fusion for Full Reference Video Quality Assessment}, 
  year={2019},
  volume={29},
  number={8},
  pages={2256-2270},
  doi={10.1109/TCSVT.2018.2868262}}

@INPROCEEDINGS{hfr_vmaf,
  author={Madhusudana, P. C. and Birkbeck, N. and Wang, Y. and Adsumilli, B. and Bovik, A. C.},
  booktitle={2021 Picture Coding Symposium (PCS)}, 
  title={High Frame Rate Video Quality Assessment using {VMAF} and Entropic Differences}, 
  year={2021},
  volume={},
  number={},
  pages={1-5},
  doi={10.1109/PCS50896.2021.9477462}}

@INPROCEEDINGS{evmaf,
  author={Zhang, F. and Katsenou, A. and Bampis, C. and Krasula, L. and Li, Z. and Bull, D.},
  booktitle={2021 Picture Coding Symposium (PCS)}, 
  title={Enhancing {VMAF} through New Feature Integration and Model Combination}, 
  year={2021},
  volume={},
  number={},
  pages={1-5},
  doi={10.1109/PCS50896.2021.9477458}}

@inproceedings{wang2003multiscale,
  author    = {Wang, Zhou and Simoncelli, Eero P. and Bovik, Alan C.},
  title     = {Multiscale Structural Similarity for Image Quality Assessment},
  booktitle = {Proceedings of the 37th Asilomar Conference on Signals, Systems and Computers},
  volume    = {2},
  pages     = {1398--1402},
  year      = {2003}
}

@article{zhang2011fsim,
  author  = {Zhang, Lin and Zhang, Lei and Mou, Xuanqin and Zhang, David},
  title   = {{FSIM}: A Feature Similarity Index for Image Quality Assessment},
  journal = {IEEE Transactions on Image Processing},
  volume  = {20},
  number  = {8},
  pages   = {2378--2386},
  year    = {2011}
}

@article{pinson2004objective,
  author  = {Pinson, Margaret H. and Wolf, Stephen},
  title   = {A New Standardized Method for Objectively Measuring Video Quality},
  journal = {IEEE Transactions on Broadcasting},
  volume  = {50},
  number  = {3},
  pages   = {312--322},
  year    = {2004}
}

@article{seshadrinathan2010motion,
  author  = {Seshadrinathan, Kalpana and Bovik, Alan C.},
  title   = {Motion Tuned Spatio-Temporal Quality Assessment of Natural Videos},
  journal = {IEEE Transactions on Image Processing},
  volume  = {19},
  number  = {2},
  pages   = {335--350},
  year    = {2010}
}

@article{wang2009mean,
  author  = {Wang, Zhou and Bovik, Alan C.},
  title   = {Mean Squared Error: Love It or Leave It? {A} New Look at Signal Fidelity Measures},
  journal = {IEEE Signal Processing Magazine},
  volume  = {26},
  number  = {1},
  pages   = {98--117},
  year    = {2009}
}

@article{chandler2013seven,
  author  = {Chandler, Damon M.},
  title   = {Seven Challenges in Image Quality Assessment: Past, Present, and Future Research},
  journal = {ISRN Signal Processing},
  volume  = {2013},
  pages   = {1--53},
  year    = {2013}
}

@ARTICLE{funque,
  author={Venkataramanan, Abhinau K. and Stejerean, Cosmin and Katsavounidis, Ioannis and Bovik, Alan C.},
  journal={IEEE Transactions on Image Processing}, 
  title={One Transform to Compute Them All: Efficient Fusion-Based Full-Reference Video Quality Assessment}, 
  year={2024},
  volume={33},
  number={},
  pages={509-524},
  doi={10.1109/TIP.2023.3345227}}

@INPROCEEDINGS{funque_hdr,
  author={Venkataramanan, Abhinau K. and Stejerean, Cosmin and Katsavounidis, Ioannis and Bovik, Alan C.},
  booktitle={2024 Picture Coding Symposium (PCS)}, 
  title={A FUNQUE Approach to the Quality Assessment of Compressed HDR Videos}, 
  year={2024},
  volume={},
  number={},
  pages={1-5},
  doi={10.1109/PCS60826.2024.10566413}}

@article{LIVEHDR2024TIP,
  author       = {Zaixi Shang and
                  Joshua Peter Ebenezer and
                  Abhinau Kumar Venkataramanan and
                  Yongjun Wu and
                  Hai Wei and
                  Sriram Sethuraman and
                  Alan C. Bovik},
  title        = {A Study of Subjective and Objective Quality Assessment of {HDR} Videos},
  journal      = {{IEEE} Trans. Image Process.},
  volume       = {33},
  pages        = {42--57},
  year         = {2024},
  doi          = {10.1109/TIP.2023.3333217},
  bibsource    = {dblp computer science bibliography, https://dblp.org}
}

@INPROCEEDINGS{CVQAHDR2025ICMEW,
  author={Sun, Wei and Cao, Linhan and Fu, Kang and Zhu, Dandan and Jia, Jun and Hu, Menghan and Min, Xiongkuo and Zhai, Guangtao},
  booktitle={2025 IEEE International Conference on Multimedia and Expo Workshops (ICMEW)}, 
  title={CompressedVQA-HDR: Generalized Full-reference and No-reference Quality Assessment Models for Compressed High Dynamic Range Videos}, 
  year={2025},
  volume={},
  number={},
  pages={1-6},
  doi={10.1109/ICMEW68306.2025.11152231}}

@article{kim2017deep,
  author  = {Kim, Jongyoo and Zeng, Hojin and Ghadiyaram, Deepti and Lee, Sanghoon and Zhang, Lei and Bovik, Alan C.},
  title   = {Deep Convolutional Neural Models for Picture-Quality Prediction: Challenges and Solutions to Data-Driven Image Quality Assessment},
  journal = {IEEE Signal Processing Magazine},
  volume  = {34},
  number  = {6},
  pages   = {130--141},
  year    = {2017}
}

@inproceedings{zhang2018unreasonable,
  author    = {Zhang, Richard and Isola, Phillip and Efros, Alexei A. and Shechtman, Eli and Wang, Oliver},
  title     = {The Unreasonable Effectiveness of Deep Features as a Perceptual Metric},
  booktitle = {Proceedings of the IEEE/CVF Conference on Computer Vision and Pattern Recognition},
  pages     = {586--595},
  year      = {2018}
}

@article{ding2020image,
  author  = {Ding, Keyan and Ma, Kede and Wang, Shiqi and Simoncelli, Eero P.},
  title   = {Image Quality Assessment: Unifying Structure and Texture Similarity},
  journal = {IEEE Transactions on Pattern Analysis and Machine Intelligence},
  volume  = {44},
  number  = {5},
  pages   = {2567--2581},
  year    = {2022}
}

@inproceedings{lao2022attentions,
  author    = {Lao, Shiqi and others},
  title     = {Attentions Help {CNNs} See Better: Attention-Based Hybrid Image Quality Assessment Network},
  booktitle = {Proceedings of the IEEE/CVF Conference on Computer Vision and Pattern Recognition Workshops},
  pages     = {1140--1149},
  year      = {2022}
}

@article{chen2024topiq,
  author  = {Chen, Jiawei and others},
  title   = {{TOPIQ}: A Top-Down Approach from Semantics to Distortions for Image Quality Assessment},
  journal = {IEEE Transactions on Image Processing},
  volume  = {33},
  pages   = {2404--2418},
  year    = {2024}
}

@inproceedings{kim2018deep,
  author    = {Kim, Wonjae and Kim, Jongyoo and Ahn, Sungjin and Kim, Jisoo and Lee, Sanghoon},
  title     = {Deep Video Quality Assessor: From Spatio-Temporal Visual Sensitivity to a Convolutional Neural Aggregation Network},
  booktitle = {European Conference on Computer Vision},
  pages     = {219--234},
  year      = {2018}
}

@inproceedings{xu2020c3dvqa,
  author    = {Xu, Jingtao and Li, Jihong and Zhou, Xiang and Zhou, Weishi and Wang, Bolun and Chen, Zhibo},
  title     = {Perceptual Quality Assessment of Internet Videos},
  booktitle = {Proceedings of the ACM International Conference on Multimedia},
  pages     = {1248--1257},
  year      = {2020}
}

@inproceedings{li2019quality,
  author    = {Li, Dingquan and Jiang, Tao and Jiang, Meng},
  title     = {Quality Assessment of In-the-Wild Videos},
  booktitle = {Proceedings of the ACM International Conference on Multimedia},
  pages     = {2351--2359},
  year      = {2019}
}

@inproceedings{starvqa,
  title     = {{StarVQA}: Space-Time Attention for Video Quality Assessment},
  author    = {Xing, Fengchuang and Wang, Yuan-Gen and Wang, Hanpin and Li, Leida and Zhu, Guopu},
  booktitle = {2022 IEEE International Conference on Image Processing (ICIP)},
  pages     = {2326--2330},
  year      = {2022},
  organization = {IEEE}
}

@article{wu2022discovqa,
  title   = {{DisCoVQA}: Temporal Distortion-Content Transformers for Video Quality Assessment},
  author  = {Wu, Haoning and Chen, Chaofeng and Liao, Liang and Hou, Jingwen and Sun, Wenxiu and Yan, Qiong and Lin, Weisi},
  eprint        = {2206.09853},
  archivePrefix = {arXiv},
  year    = {2022}
}

@article{ma2017dipiq,
   title={dipIQ: Blind Image Quality Assessment by Learning-to-Rank Discriminable Image Pairs},
   volume={26},
   ISSN={1941-0042},
   DOI={10.1109/tip.2017.2708503},
   number={8},
   journal={IEEE Transactions on Image Processing},
   publisher={Institute of Electrical and Electronics Engineers (IEEE)},
   author={Ma, Kede and Liu, Wentao and Liu, Tongliang and Wang, Zhou and Tao, Dacheng},
   year={2017},
  pages={3951–3964} }

@inproceedings{he2016deep,
  author    = {He, Kaiming and Zhang, Xiangyu and Ren, Shaoqing and Sun, Jian},
  title     = {Deep Residual Learning for Image Recognition},
  booktitle = {Proceedings of the IEEE Conference on Computer Vision and Pattern Recognition},
  pages     = {770--778},
  year      = {2016}
}

@article{russakovsky2015imagenet,
  title={Imagenet large scale visual recognition challenge},
  author={Russakovsky, Olga and Deng, Jia and Su, Hao and Krause, Jonathan and Satheesh, Sanjeev and Ma, Sean and Huang, Zhiheng and Karpathy, Andrej and Khosla, Aditya and Bernstein, Michael and others},
  journal={International journal of computer vision},
  volume={115},
  number={3},
  pages={211--252},
  year={2015},
  publisher={Springer}
}

@article{wu2023qalign,
  title={{Q-Align}: Teaching LMMs for Visual Scoring via Discrete Text-Defined Levels},
  author={Wu, Haoning and Zhang, Zicheng and Zhang, Weixia and Chen, Chaofeng and Li, Chunyi and Liao, Liang and Wang, Annan and Zhang, Erli and Sun, Wenxiu and Yan, Qiong and Min, Xiongkuo and Zhai, Guangtao and Lin, Weisi},
  journal={arXiv preprint arXiv:2312.17090},
  year={2023},
  institution={Nanyang Technological University and Shanghai Jiao Tong University and Sensetime Research},
  note={Equal Contribution by Wu, Haoning and Zhang, Zicheng. Corresponding Authors: Zhai, Guangtao and Lin, Weisi.}
}

@misc{ge2024lmmvqaadvancingvideoquality,
      title={LMM-VQA: Advancing Video Quality Assessment with Large Multimodal Models}, 
      author={Qihang Ge and Wei Sun and Yu Zhang and Yunhao Li and Zhongpeng Ji and Fengyu Sun and Shangling Jui and Xiongkuo Min and Guangtao Zhai},
      year={2024},
      eprint={2408.14008},
      archivePrefix={arXiv},
      primaryClass={cs.CV},
      url={https://arxiv.org/abs/2408.14008}, 
}

@inproceedings{dai2017deformable,
  author    = {Dai, Jifeng and Qi, Haozhi and Xiong, Yuwen and Li, Yi and Zhang, Guodong and Hu, Han and Wei, Yichen},
  title     = {Deformable Convolutional Networks},
  booktitle = {Proceedings of the IEEE International Conference on Computer Vision},
  pages     = {764--773},
  year      = {2017}
}

@inproceedings{woo2018cbam,
  author    = {Woo, Sanghyun and Park, Jongchan and Lee, Joon-Young and Kweon, In So},
  title     = {{CBAM}: Convolutional Block Attention Module},
  booktitle = {European Conference on Computer Vision},
  pages     = {3--19},
  year      = {2018}
}

@misc{Majeedi2023FullRV,
  title={Full Reference Video Quality Assessment for Machine Learning-Based Video Codecs},
  author={Abrar Majeedi and Babak Naderi and Yasaman Hosseinkashi and Juhee Cho and Ruben Alvarez Martinez and Ross Cutler},
  eprint={2309.00769},
  year={2023},
  archivePrefix={arXiv},
}

@inproceedings{herb2025evaluating,
  author    = {Benjamin Herb and Rakesh Rao Ramachandra Rao and Steve G{\"o}ring and Alexander Raake},
  title     = {Evaluating Video Quality Metrics for Neural and Traditional Codecs using 4K/UHD-1 Videos},
  booktitle = {Proc. Picture Coding Symposium (PCS)},
  year      = {2025}
}

@inproceedings{jenadeleh2025jpegai,
  author    = {Mohsen Jenadeleh and Jon Sneyers and Panqi Jia and Shima Mohammadi and Jo{\~a}o Ascenso and Dietmar Saupe},
  title     = {Subjective Visual Quality Assessment for High-Fidelity Learning-Based Image Compression},
  booktitle = {Proc. QoMEX},
  year      = {2025}
}

@ARTICLE{CONVIQT,
  author={Madhusudana, Pavan C. and Birkbeck, Neil and Wang, Yilin and Adsumilli, Balu and Bovik, Alan C.},
  journal={IEEE Transactions on Image Processing}, 
  title={CONVIQT: Contrastive Video Quality Estimator}, 
  year={2023},
  volume={32},
  number={},
  pages={5138-5152},
  doi={10.1109/TIP.2023.3310344}}

@conference{Segall2023JVETAC2011,
  author       = {Segall, A. and François, E. and Husak, W. and Iwamura, S. and Rusanovskyy, D.},
  title        = {{VTM} and {HM} Common Test Conditions and Evaluation Procedures for {HDR/WCG} Video},
  organization  = {Joint Video Experts Team (JVET) of ITU-T SG 16 WP 3 and ISO/IEC JTC 1/SC 29 JVET-AC2011},
  year         = {2023},
}

@INPROCEEDINGS{pu21,
  author={Mantiuk, Rafa∤ K. and Azimi, Maryam},
  booktitle={2021 Picture Coding Symposium (PCS)}, 
  title={PU21: A novel perceptually uniform encoding for adapting existing quality metrics for HDR}, 
  year={2021},
  volume={},
  number={},
  pages={1-5},
  doi={10.1109/PCS50896.2021.9477471}}

@misc{hdrvdp32023arXiv,
      title={HDR-VDP-3: A multi-metric for predicting image differences, quality and contrast distortions in high dynamic range and regular content}, 
      author={Rafal K. Mantiuk and Dounia Hammou and Param Hanji},
      year={2023},
      eprint={2304.13625},
      archivePrefix={arXiv},
      primaryClass={eess.IV},
      url={https://arxiv.org/abs/2304.13625}, 
}

@article{mantiuk2011hdr,
  author  = {Mantiuk, Rafa{\l} and Kim, K. J. and Rempel, Allan G. and Heidrich, Wolfgang},
  title   = {{HDR-VDP-2}: A Calibrated Visual Metric for Visibility and Quality Predictions in All Luminance Conditions},
  journal = {ACM Transactions on Graphics},
  volume  = {30},
  number  = {4},
  pages   = {1--14},
  year    = {2011}
}

@article{colorvideovdp,
  author  = {Mantiuk, Rafa{\l} K. and others},
  title   = {{ColorVideoVDP}: A Visual Difference Predictor for Image, Video and Display Distortions},
  journal = {ACM Transactions on Graphics},
  volume  = {40},
  number  = {4},
  year    = {2021}
}

@misc{chubarau2024adaptingpretrainednetworksimage,
      title={Adapting Pretrained Networks for Image Quality Assessment on High Dynamic Range Displays}, 
      author={Andrei Chubarau and Hyunjin Yoo and Tara Akhavan and James Clark},
      year={2024},
      eprint={2405.00670},
      archivePrefix={arXiv},
      primaryClass={cs.CV},
}

@InProceedings{Saini_2024_WACV,
    author    = {Saini, Shreshth and Saha, Avinab and Bovik, Alan C.},
    title     = {{HIDRO-VQA}: High Dynamic Range Oracle for Video Quality Assessment},
    booktitle = {Proceedings of the IEEE/CVF Winter Conference on Applications of Computer Vision (WACV) Workshops},
    month     = {January},
    year      = {2024},
    pages     = {469-479}
}

@inproceedings{feng2024rankdvqa,
  title     = {{RankDVQA}: Deep {VQA} Based on Ranking-Inspired Hybrid Training},
  author    = {Feng, Chen and Danier, Duolikun and Zhang, Fan and Bull, David},
  booktitle = {Proceedings of the IEEE/CVF Winter Conference on Applications of Computer Vision},
  pages     = {1648--1658},
  year      = {2024}
}

@misc{TVQA,
    title = {Call for Technical Proposals for the AVS Objective Video Quality Assessment Standard},
    organization = {AVS N3856},
    year = {2024},
}

@misc{CDVL,
  title  = {{Consumer Digital Video Library}},
  author = {{Institute for Telecommunication Sciences}},
  howpublished = {\url{https://www.cdvl.org/}},
}

@misc{Xiph,
  title  = {{Xiph Video Codec Test Media}},
  author = {{Xiph.Org Foundation}},
  howpublished = {\url{https://media.xiph.org/}},
}

@misc{IVP,
  title  = {{IVP Subjective Quality Video Database}},
  author = {{Chinese University of Hong Kong}},
  howpublished = {\url{https://ivp.ee.cuhk.edu.hk/research/database/subjective/index.shtml}},
}

@misc{LIVE-NRLX-II,
author = {Bampis, Christos and Li, Zhi and Katsavounidis, Ioannis and Huang, Te-Yuan and Ekanadham, Chaitanya and Bovik, Alan},
year = {2018},
title = {Towards Perceptually Optimized End-to-end Adaptive Video Streaming},
eprint = {1808.03898},
archivePrefix = {arXiv},
}

@INPROCEEDINGS{GamingVideoSET,
  author={Barman, Nabajeet and Zadtootaghaj, Saman and Schmidt, Steven and Martini, Maria G. and Möller, Sebastian},
  booktitle={2018 16th Annual Workshop on Network and Systems Support for Games (NetGames)}, 
  title={GamingVideoSET: A Dataset for Gaming Video Streaming Applications}, 
  year={2018},
  volume={},
  number={},
  pages={1-6},
  doi={10.1109/NetGames.2018.8463362}}

@article{mapping_PLCC,
  author       = {Hamid R. Sheikh and
                  Muhammad F. Sabir and
                  Alan C. Bovik},
  title        = {A Statistical Evaluation of Recent Full Reference Image Quality Assessment
                  Algorithms},
  journal      = {{IEEE} Trans. Image Process.},
  volume       = {15},
  number       = {11},
  pages        = {3440--3451},
  year         = {2006},
  doi          = {10.1109/TIP.2006.881959},
  bibsource    = {dblp computer science bibliography, https://dblp.org}
}

@InProceedings{Blau_2018_CVPR,
author = {Blau, Yochai and Michaeli, Tomer},
title = {The Perception-Distortion Tradeoff},
booktitle = {Proceedings of the IEEE Conference on Computer Vision and Pattern Recognition (CVPR)},
month = {June},
year = {2018}
}

\end{document}